\documentclass{article}

\PassOptionsToPackage{numbers, sort&compress}{natbib}
\usepackage[final]{neurips_2024}

\usepackage{makecell}
\usepackage{graphicx}
\usepackage[utf8]{inputenc} % allow utf-8 input
\usepackage[T1]{fontenc}    % use 8-bit T1 fonts
\usepackage{hyperref}       % hyperlinks
\hypersetup{
  hypertexnames=false
}
\usepackage{url}            % simple URL typesetting
\usepackage{booktabs}       % professional-quality tables
\usepackage{amsfonts}       % blackboard math symbols
\usepackage{nicefrac}       % compact symbols for 1/2, etc.
\usepackage{microtype}      % microtypography
\usepackage{xcolor}         % colors

\usepackage{subfigure}
\usepackage{graphicx}
\usepackage{multirow}
\newcommand{\paratitle}[1]{\vspace{1.5ex}\noindent\textbf{#1}}
\newcommand\Vector{\bm}
\newcommand\Matrix{\mathbf}            % \matrix is defined in LaTeX2e kernel
\newcommand\Tensor{\mathcal}
\usepackage{colortbl}
\usepackage{color, xcolor}
\usepackage{amsmath}

\usepackage{algpseudocode}
\usepackage[toc,page]{appendix}
\usepackage{setspace}
\usepackage{algorithm}
\usepackage{caption}
\usepackage{amsthm}
\usepackage{tikz}
\newtheorem{thm}{\bf Theorem}

\usepackage{wrapfig}
\usepackage{amssymb}
\usepackage{amsmath} 
\usepackage{booktabs}
\usepackage{enumerate}
\usepackage{graphicx}
\usepackage{subfigure}
\usepackage{xspace}
\usepackage{float}
\usepackage{bbm}
\usepackage{bm}
\usepackage{multirow}
\usepackage{booktabs}
\usepackage{color}
\usepackage{framed}
\usepackage{stfloats}
\usepackage{iitem}
\usepackage[T1]{fontenc}
\definecolor{shadecolor}{RGB}{180,180,180}
\usepackage{url}
\newcommand{\ie}{\emph{i.e.,}\xspace}

\newcommand{\eg}{\emph{e.g.,}\xspace}

\newcommand{\ignore}[1]{}

\usepackage{algpseudocode}
\usepackage{amsthm}
\usepackage{amsmath}
\usepackage{tikz}

\usepackage{amsmath}
\usepackage{colortbl}
\usepackage{color, xcolor}
\usepackage{graphicx}

\title{Over-parameterized Student Model via Tensor Decomposition
Boosted Knowledge Distillation}

\author{%
  Yu-Liang Zhan \\
  Gaoling School of Artificial Intelligence\\
  Renmin University of China \\
  \texttt{zhanyuliang@ruc.edu.cn} \\\And
  Zhong-Yi Lu \\
  School of Physics\\
  Renmin University of China \\
  \texttt{zlu@ruc.edu.cn} \\\AND
  Hao Sun\thanks{Corresponding authors.} \\
  Gaoling School of Artificial Intelligence\\
  Renmin University of China \\
  \texttt{haosun@ruc.edu.cn} \\\And
  Ze-Feng Gao\footnotemark[1] \\
  School of Physics\\
  Renmin University of China \\
  \texttt{zfgao@ruc.edu.cn} \\
}

\begin{document}
\maketitle

\begin{abstract}
Increased training parameters have enabled large pre-trained models to excel in various downstream tasks. Nevertheless, the extensive computational requirements associated with these models hinder their widespread adoption within the community. We focus on Knowledge Distillation (KD), where a compact student model is trained to mimic a larger teacher model, facilitating the transfer of knowledge of large models. In contrast to much of the previous work, we scale up the parameters of the student model during training, to benefit from over-parameterization without increasing the inference latency. In particular, we propose a tensor decomposition strategy that effectively over-parameterizes the relatively small student model through an efficient and nearly lossless decomposition of its parameter matrices into higher-dimensional tensors. To ensure efficiency, we further introduce a tensor constraint loss to align the high-dimensional tensors between the student and teacher models. Comprehensive experiments validate the significant performance enhancement by our approach in various KD tasks, covering computer vision and natural language processing areas. Our code is available at~\url{https://github.com/intell-sci-comput/OPDF}.
%All the source codes will be released after the review period.

\end{abstract}
\section{Introduction}
\label{sec-intro}
Large-scale pre-trained models are gradually achieving remarkable milestones due to the exhibit of remarkable performance across various tasks~\cite{Kaplan2020ScalingLF, devlin2018bert,liu2019roberta, raffel2020exploring, gu2023eva2, wang2023large, Dosovitskiy2021VisionTransformer}. These models leverage extensive pre-training data and parameters, enabling them to effectively encapsulate a significant breadth of world knowledge~\cite{roberts2020much, jiang2020can} and exhibit strong generalization capabilities across diverse tasks~\cite{Kaplan2020ScalingLF, hoffmannTrainingComputeOptimalLarge2022, alabdulmohsinGettingViTShape2023,brown2020language,lester2021power}. Following this trajectory, the utilization of increased data and parameters has emerged as a notable trend in enhancing the performance of pre-trained models in recent years, leading to the number expansion of pre-trained model parameters from millions to billions~\cite{raffel2020exploring,chowdhery2022palm,chen2022pali}.

Despite their impressive performance, the substantial storage demands and high computational complexity hinder the practical deployment of these models in real-world applications. Therefore, on the one hand, some studies focus on pre-training relatively smaller models~(such as BERT-base-uncased~\cite{devlin2018bert}) on domain-specific or task-specific corpora~\cite{gururangan2020don,chang2020pre,zhang2020dialogpt}. However, due to the lesser over-parameterization of small models compared to large ones, their generalization capability often falls short, resulting in suboptimal fine-tuning performance on downstream tasks. On the other hand, model compression methods, such as pruning less informative parameters~\cite{anwar2017structured, malach2020proving, zhang2021validating} or utilizing \emph{knowledge distillation}~(KD)~\cite{hintonDistillingKnowledgeNeural2015} to transfer knowledge from larger models (teachers) to smaller ones (students), have been proposed. KD has swiftly diversified into numerous branches, primarily falling into two categories: \ie logits-based~\cite{hintonDistillingKnowledgeNeural2015, jiaoTinyBERTDistillingBERT2020, xuImprovingBERTFineTuning2020, zhaoDecoupledKnowledgeDistillation2022, huangKnowledgeDistillationStronger2022} and features-based~\cite{romeroFitNetsHintsThin2015, parkRelationalKnowledgeDistillation2019, chenDistillingKnowledgeKnowledge2021, wangMiniLMv2MultiHeadSelfAttention2021} depending on the source of student model knowledge. 
Nevertheless, as student models have fewer trainable parameters and limited capacity, a significant performance gap remains between student and teacher models.

To address the disparity between small and large models, this study aims to over-parameterize small student models as large ones during distillation training to enhance their generalization capability. Typically, most parameters of student models are stored as matrices. Through tensor decomposition techniques~\cite{tucker1966some,henry19928,oseledets2011tensor,gao2020compressing}~(\eg Singular Value Decomposition), each matrix can be factorized into a set of matrices, effectively increasing the total number of parameters during distillation. Moreover, after convergence, the factorized matrices can be merged to reorganize the parameter matrix of the student model. This paradigm leverages the benefits of over-parameterization during training without increasing the inference latency of student models.

However, incorporating tensor decomposition into over-parameterizing student models poses two major concerns that must be addressed. First, the potential information loss caused by tensor decomposition should be minimized, as small computation errors may accumulate and propagate exponentially within the stacked layers of student models. Second, in the over-parameterized student models, there is no effective mechanism to ensure the consistency of information between student and teacher models. Therefore, it is essential to choose appropriate tensor decomposition methods and design loss functions for high-order tensors to ensure the effective transfer of information from teacher to student models.

To address the above issues, we introduce the matrix product operator (MPO)~\cite{gao2020compressing} technique as the tensor decomposition strategy. The MPO decomposition, widely used in quantum many-body physics, efficiently factorizes any matrix with arbitrary dimensions into a set of higher-dimensional tensors, which can reconstruct the original matrix in almost lossless conditions~\cite{gao2020compressing,liu2021enabling,gao2022parameter,gao-etal-2023-small}. These advantages make MPO an ideal method for over-parameterizing student models during distillation. Based on MPO, we also devise high-order tensor alignment losses for student and teacher models to ensure the effective transfer of information in tensor representation.

Therefore, in this paper, we propose a general \underline{O}ver-\underline{P}arameterization \underline{D}istillation \underline{F}ramework, namely \textbf{OPDF}, to improve the performance of knowledge distillation. Given the parameter matrices of a student model, we first over-parameterize them through MPO decomposition and then utilize high-order tensor alignment losses to ensure efficient information transfer. This framework only modifies the distillation training process, making it applicable to various student models and natural language processing~(NLP) and computer vision~(CV) tasks. We conduct extensive experiments in both NLP and CV domains. Experimental results demonstrate that our OPDF significantly enhances the effectiveness of model distillation, \eg improving BERT-base KD +1.6 on average. Moreover, our approach also enables the student model to achieve performance nearly on par with the teacher model, \eg AD-KD+Ours~(83.4) \emph{v.s.} BERT-base~(83.4) in average metric on GLUE.

\section{Related work}
\label{sec-related-work}
\paragraph{Large Scale Pre-trained Models} 
Large-scale pre-trained models have achieved remarkable success in many fields~(\eg natural language processing~(NLP)~\cite{touvron2023llama} and computer vision~(CV)~\cite{chen2022pali, tian2019contrastive}). Since the introduction of the Transformer architecture~\cite{vaswani2017attention}, the pre-training and fine-tuning paradigm in NLP, exemplified by models like BERT~\cite{devlin2018bert} and T5~\cite{raffel2020exploring}, has shown outstanding performance across multiple tasks. Furthermore, the emergence of models like GPT-3 has demonstrated that increasing model size can significantly improve performance on low-resource tasks~\cite{brown2020language}. In the field of computer vision, models based on Transformers, such as ViT~\cite{Dosovitskiy2021VisionTransformer}, have also performed exceptionally well. In our research, we improve the distillation process by increasing the parameters during the training phase of the student model, without introducing additional inference latency to the student model.

\paragraph{Knowledge Distillation}                    
Knowledge Distillation~(KD) methods are commonly used to compress models by transferring knowledge from a larger \emph{teacher model} to a smaller \emph{student model}. 
Building upon the initiative work by ~\cite{hintonDistillingKnowledgeNeural2015}, the researchers have exploited the logits follows up with different techniques in the computer vision field, \eg minimizing KL-Divergence~(DKD~\cite{zhaoDecoupledKnowledgeDistillation2022}) or a Pearson correlation~(DIST~\cite{huangKnowledgeDistillationStronger2022}). 
Logit-based methods have been also popular in NLP~\cite{jiao2019tinybert,touvron2021training}. Features-based methods have tried to align the features from intermediate layers of teacher and student models and minimize the differences~\cite{gou2021knowledge}. After the intermediate representations have been introduced~\cite{romeroFitNetsHintsThin2015}, a mount of features-based KD methods have been proposed to match the features, such as LGTM~\cite{ren2023tailoring}, DBKD~\cite{zhou2023bridging} and AD-KD~\cite{wu2023ad}. 
However, the capacity gap between the teacher and student models makes it difficult to imitate the hidden representations of the teacher~\cite{liang2023less}.
Different from these existing KD methods, our proposed OPDF has utilized MPO decomposition to over-parameterize the student model in the training procedure to improve the student model generalization capability, which can minimize the capacity gap efficiently. 

\paragraph{Matrix Product Operators} 
Matrix Product Operators~(MPOs)~\cite{pirvu2010matrix,gao2020compressing}, also known as tensor-train operators~(TTOs)~\cite{oseledets2011tensor}, have been proposed for a more efficient representation of the linear structure of neural networks~\cite{gao2020compressinglstm,sun2020model}. A large number of typical applications have utilized MPO-based methods to compress linear layers~\cite{novikov2015tensorizing} and convolutional kernels~\cite{garipov2016ultimate} in the parameter matrices of deep models. Furthermore, existing works have applied the MPO method for lightweight fine-tuning of ALBERT~\cite{liu2021enabling}, the efficient expansion for the MoE framework~\cite{gao2022parameter}, the over-parameterization tuning process for PLMs~\cite{gao-etal-2023-small}, construct efficient PLM architecture~\cite{liu2023enhancing,liu2024unlocking} and compressing datasets~\cite{gao2024compression}. Unlike existing methods, our approach focuses on utilizing MPO decomposition to map parameters from low-dimensional spaces to high-dimensional spaces, to over-parameterize the student model during the distillation process, allowing the student model to benefit from more parameters and achieve better distillation results.

\section{Preliminary}
\paragraph{Tensor Product}
We denote a tensor $\Tensor{T}_{i_1,i_2,\dots, i_n}$ as an array with $n$ indices, where $\{i_1, i_2,\dots, i_n\}$ denotes the dimensions of the $n$ indices, respectively. In this manner, a vector~(\ie $\Vector{v}$) can be considered a $1$-order tensor, while a matrix~(\ie $\Matrix{W}$) can be regarded as a $2$-order tensor. Consider ${\psi_1, \dots, \psi_{p}}$ and ${\phi_1, \dots, \phi_{q}}$ as the orthonormal bases of tensors $\Tensor{T}^{(1)}$ and $\Tensor{T}^{(2)}$, respectively. The tensor product, denoted as $\otimes$, can be obtained through the contraction of $\Tensor{T}^{(1)}$ and $\Tensor{T}^{(2)}$. Formally, the tensor contraction of  $\Tensor{T}^{(1)} = \sum_{i=1}^{p} a_i\psi_{i_1}$ and $\Tensor{T}^{(2)} = \sum_{j=1}^{q} b_j\phi_{i_2}$ is defined as follow:
\begin{equation}
    \Tensor{T}^{(1)} \otimes \Tensor{T}^{(2)} =  \sum_{i=1}^{p} \sum_{j=1}^{q}a_i b_j \psi_{i_1}\otimes \phi_{i_2}.
\end{equation}
The set ${\psi_{i_1} \otimes \phi_{i_2}}$ constitutes the orthonormal basis of the resulting vector Hilbert space, with the dimensionality of this Hilbert space being the product~(i.e., $p\times q$) of $\Tensor{T}^{(1)}$ and $\Tensor{T}^{(2)}$.

\paragraph{Tensor Decomposition}
Tensor decomposition can be viewed as the reverse operation of the tensor product. A commonly employed approach is the singular value decomposition~(SVD) algorithm. Given a tensor $\Tensor{T} \in \mathbb{R}^{i_1 \times \cdots \times i_n}$, the SVD operation performed $n$ times can decompose this tensor into $n$ local tensors ${\Tensor{T}^{(k)}}_{k=1}^{n}$. Conversely, the decomposed tensors can reconstruct the original tensor by sequentially applying the tensor product operator.

\section{Method}
\label{method}
In this section, we describe our proposed over-parameterized distillation framework. We first outline our approach, then introduce the details of matrix product operator decomposition and the over-parameterized student model strategy, and finally present the tensor alignment loss.

\subsection{Overview}
\begin{figure}[t!]
    \centering
    \small
    \includegraphics[width=0.85\linewidth]{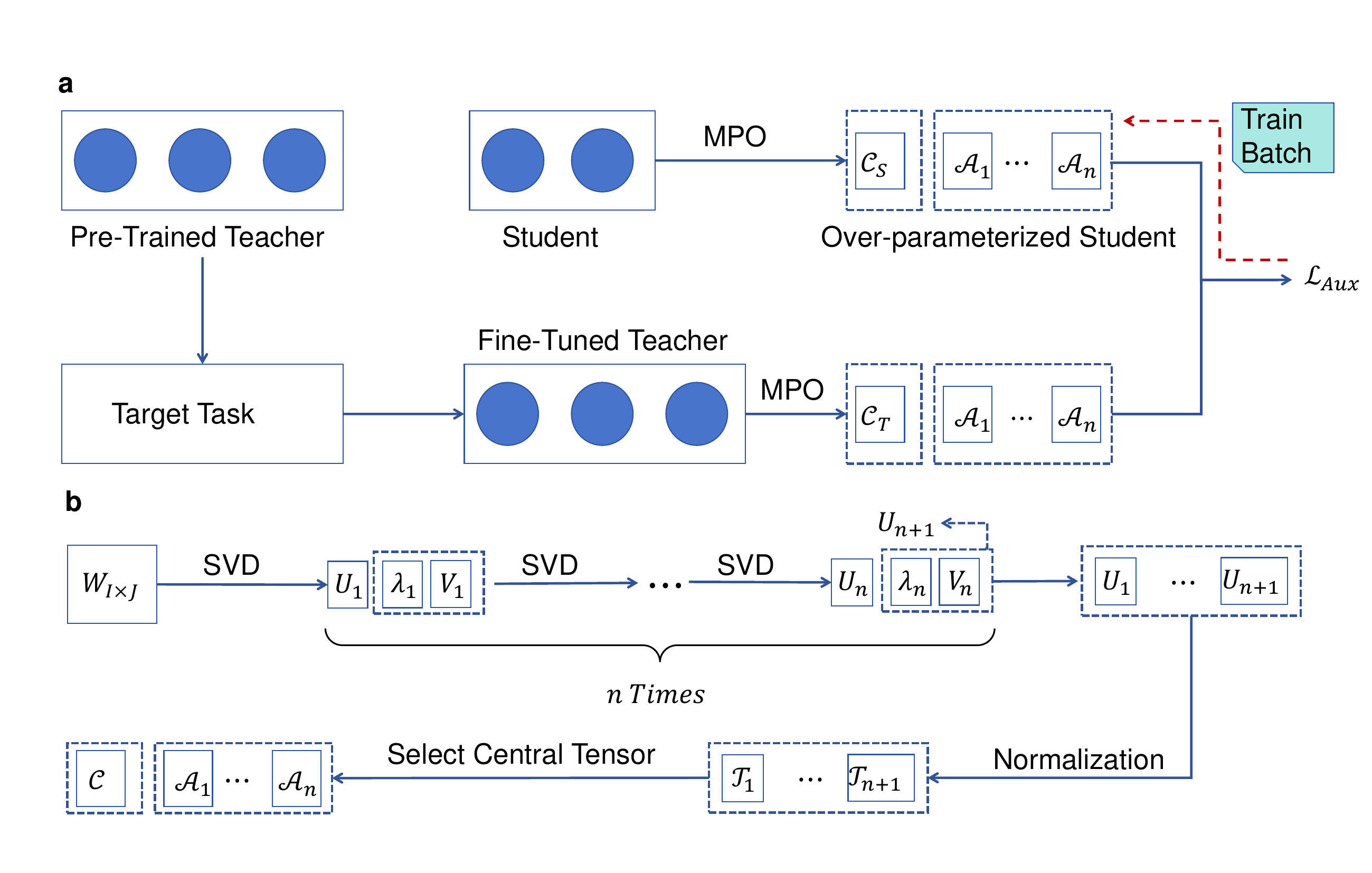}
    \vspace{-6pt}
    \caption{The overview of over-parameter distillation framework~(OPDF) for knowledge ditillation.
    {\bf{a}}, We use MPO decomposition to realize the over-parameter procedure for the student model. The auxiliary tensors of the student model are trained to imitate the auxiliary tensors of the teacher model closely.
    {\bf{b}}, We present an illustrative example of MPO decomposition. A parameter matrix $\mathbf{W}_{I\times J}$ is decomposed into central tensor and auxiliary tensors.
    }
    \label{fig-main}
\end{figure}
Current distillation methods primarily enhance the performance of student models by introducing constraints on logits or features between the student and teacher models. In contrast to these methods, our approach not only utilizes tensor decomposition to over-parameterize the student model for performance improvement but also designs alignment loss functions for the decomposed high-order tensors to further enhance the performance of the student model. To achieve this goal, we employ a tensor decomposition method to decompose the parameter matrices of the teacher and student models into a series of high-order tensor products. These high-order tensors can be used to reconstruct the original parameter matrices while significantly increasing the number of trainable parameters in the student model. After reconstruction, the student model has the same number of parameters as the original matrix without increasing inference time and model size. Additionally, by introducing distillation loss functions to allow the student model to learn from the teacher model in tensor representation, the effectiveness of knowledge distillation is further enhanced.

In our proposed over-parameterized distillation framework, we integrate a tensor decomposition strategy based on MPO into the student model to enlarge the parameter matrix~(Section~\ref{subsec-over-paramterization}). Furthermore, we design a tensor alignment loss function to enhance the performance of the student model in the context of knowledge distillation~(Section~\ref{subsec-assisted loss}). An overview of our approach is depicted in Figure~\ref{fig-main}. We also provide a detailed description of our over-parameterized distillation framework in Algorithm~\ref{alg:over-all-alg}.

\subsection{Over-paramterization Distillation Framework via MPO Decomposition}
\label{subsec-over-paramterization}

To leverage the advantages of over-parameterization during knowledge distillation, our method utilizes the MPO, a tensor decomposition technique that increases the number of model parameters. In this part, we initially present the specifics of the MPO method and subsequently outline its adaptation for over-parameterizing the student model.

\paragraph{Matrix Product Operator Decomposition}
The MPO decomposition is an efficient algorithm capable of factorizing a parameter matrix $\Matrix{W} \in \mathcal{R}^{I\times J}$ into a sequential product of multiple tensors~\cite{gao2020compressing}. Formally, given a matrix $\Matrix{M}\in \mathbb{R}^{I\times J}$, its MPO decomposition into a product of $n$ local tensors can be represented as:
\begin{equation}
    \textsc{MPO}~(\Matrix{M})=\prod_{k=1}^{n} \Tensor{T}_{(k)}[d_{k-1},i_k,j_k,d_k].
    \label{eq:mpo}
\end{equation}
The tensor $\Tensor{T}{(k)}[d_{k-1},i_k,j_k,d_k]$ is a 4th-order tensor with dimensions $d_{k-1}\times i_k \times j_k \times d_k$, where $\prod_{k=1}^{n}i_k=I, \prod_{k=1}^{n}j_k=J$, and $d_0=d_n=1$. To link two sequence tensors, we have adopted the concept of a \emph{bond} following the work of~\cite{pirvu2010matrix}. The bond dimension $d_k$ is defined by:

\begin{equation}
    d_k = \min\bigg(\prod_{m=1}^k i_m\times j_m, \prod_{m=k+1}^n i_m\times j_m\bigg).
    \label{eq:d-k}
\end{equation}
From Eq.~\eqref{eq:d-k}, we can see that $d_k$ will be large in the middle and small on both sides. Following~\cite{liu2021enabling}, we refer to the tensor right in the middle as \emph{central tensor}, and the rest as \emph{auxiliary tensor}. Figure~\ref{fig-main}(b) presents the illustration of MPO decomposition. You can find additional descriptions of tensors and MPO in Appendix~\ref{appendix_a}.

\paragraph{Over-parameterzing Student Model.}
Utilizing the MPO method within the framework of knowledge distillation, our objective is to extend the parameter scale of the student model, capitalizing on over-parameterization. More specifically, we can employ the MPO method to break down a portion of the parameter matrices into multiple tensors as illustrated in Eq.~\eqref{eq:mpo}. Following MPO decomposition, the parameter count of the matrix $\Matrix{W}$ will increase based on the values of $\{d_k\}_{k=1}^{m}$, $\{i_k\}_{k=1}^{m}$, and $\{j_k\}_{k=1}^{m}$. The precise augmentation in parameter count, denoted as $N_{add}$, can be computed as follows:
\begin{equation}
    N_{add} = \sum_{k=1}^{m} i_kj_kd_{k-1}d_k - \prod_{k=1}^{m}i_kj_k.
    \label{eq:parameter-add}
\end{equation}
Therefore, during the knowledge distillation procedure, we can adopt MPO on student model parameter matrices to generate their corresponding multiple tensors. In this way, we can scale up the total parameter of the number of the student model without increasing its inference time consumption. After training the over-parameterized student model to convergence, we will perform tensor contraction on these decomposed tensors, to reconstruct the parameter matrices of the student model in almost lossless conditions which is detailed in Appendix~\ref{appendix_b}. This new student model has the same parameter number and inference latency as the original one and has benefited from over-parameterization during training.

\subsection{Assisted Constraints for Knowledge Distillation}
\label{subsec-assisted loss}
\paragraph{Revisiting Prediction Match of Knowledge Distillation}
Traditional knowledge distillation involves two stages: fine-tuning the teacher model for a specific task, followed by training strategies to constrain the student model to closely approximate the teacher model. These processes aim to transfer the knowledge from the teacher to the student model. Recent studies have mainly focused on directly learning from the features and logits of the teacher model to transfer crucial knowledge~\cite{xu2020bert, jiaoTinyBERTDistillingBERT2020}.

However, these methods are limited by the capacity of the student model due to the limitation of total parameters. Moreover, this distillation approach based on cross-entropy loss constraints may lead to the student model \emph{losing its ability to learn independently}. We aim to design a novel model distillation framework to enable the student model not only to effectively learn the knowledge from the teacher model but also to maintain its ability to learn independently.

\paragraph{Distillation Loss for Auxilary Tensors.}
To achieve the goal of "learning knowledge from the teacher model while maintaining the ability of the student model to learn independently," we introduce a high-order tensor alignment training method based on the MPO decomposition. A crucial merit of MPO decomposition is its ability to reorganize and aggregate the core information, decomposing the weight matrices into a central tensor~(containing a large number of parameters and important information) and auxiliary tensors (containing fewer parameters and additional information to the central tensor)~\cite{gao2022parameter, liu2021enabling}. Therefore, in the knowledge distillation, in addition to minimizing the cross-entropy loss concerning the ground truth, we add a loss constraint for aligning the auxiliary tensors between the student and teacher models:
\begin{equation}
    \mathcal{L}_{Aux}=\frac{1}{n}\sum_{k=1}^n\textsc{MSE}~(\mathcal{A}_{s,k},\mathcal{A}_{t,k}),
    \label{eq:loss_auxiliary}
\end{equation}
where the matrices $\mathcal{A}_{t,k}$ and $\mathcal{A}_{s,k}$ refer to the auxiliary tensor of student and teacher models with the same dimensions respectively. \textsc{MSE} means the mean-square error loss function. To ensure that the student model learns from the teacher while preserving its central tensor for independent learning, we minimize the mean-square error loss between the auxiliary tensors of both the student and teacher models. Since this distillation framework is based on improvements to the weight matrices, it is orthogonal to most current distillation methods. Therefore, it can further enhance the distillation effectiveness based on existing distillation methods~(as thoroughly discussed in the experimental section). Hence, it can be widely applied to various knowledge distillation models.

\section{Experiments}
\label{experiments}
In this section, we assess the efficacy of our approach within two renowned domains: computer vision and natural language processing. Notably, the OPDF is designed to complement existing distillation techniques. Consequently, we apply our proposed OPDF with various standard distillation methods to validate its effectiveness. In the subsequent section, we detail our experimental setup's datasets and baseline methods. We then present the primary results achieved with the OPDF and provide a thorough analysis. Furthermore, we examine the influence of the degree of over-parameterization, MPO strategy and the learning rate on the performance of OPDF. We report the memory and time cost of experiments in Appendix~\ref{cost}.

\subsection{Experimental Setup}
\paragraph{Datasets and Metrics}
For NLP tasks, we evaluate our approach on text classification tasks in GLUE benchmark~\cite{wang2018glue}. The tasks encompassed in our evaluation include RTE, MRPC, STS-B, CoLA, SST-2, QNLI, QQP, and MNLI. To facilitate comparison with baselines, we employ the F1 score and accuracy as metrics for MRPC and QQP, Matthew's correlation coefficient for CoLA, and the average of Pearson and Spearman correlations for STS-B. Accuracy is used as the metric for the remaining tasks, with the result for MNLI reported as the average across the matched~(MNLI-m) and mismatched~(MNLI-mm) domains. Additionally, we calculate the average score across all tasks to provide a comprehensive performance measure. In the context of CV tasks, we have applied the OPDF to the distillation of Vision Transformers~(ViT) for image classification~\cite{Dosovitskiy2021VisionTransformer}. This was done using the ImageNet-21k dataset~\cite{deng2009imagenet}, ImageNet-1k, ImageNet Real~\cite{beyer2020we}, and ImageNet V2~\cite{recht2019imagenet} datasets. For these datasets, we use accuracy as the primary evaluation metric. 

\paragraph{Baseline Methods}
For NLP tasks, we implement OPDF on previous KD methods: BERT-of-Theseus~\cite{xu2020bert}, LGTM~\cite{ren2023tailoring}, DBKD~\cite{zhou2023bridging} and AD-KD~\cite{wu2023ad}. We replicated the baselines using the publicly released code to assess their performance on the test set. Additionally, LGTM was not previously evaluated across all tasks in its original publications, and we have addressed this omission using the provided code. It is important to note that DBKD is designed to estimate logits from decision distributions~\cite{zhou2023bridging}, and therefore we do not report performance on the STS-B task. For all experiments in natural language processing, we demonstrate the effectiveness of our method during the fine-tuning stage. We implement the teacher model as the fine-tuned ``BERT-base-uncased'' model~\cite{devlin2018bert}. 
In the context of CV tasks, TinyViT~\cite{wu2022tinyvit}, which introduces a rapid pre-training framework, has emerged as a classical distillation method for ViT. The original paper on TinyViT discusses three versions of the model with varying parameter counts: TinyViT-5M, TinyViT-11M, and TinyViT-21M. To incorporate high-order tensor alignment loss into the distillation phase, we utilize CLIP-VIT-L/14~\cite{radford2021learning, Dosovitskiy2021VisionTransformer}, a variant of ViT, as the teacher model in our experiments. To assess the efficacy of OPDF, we pretrain the distillation model on ImageNet-21k and evaluate its linear probe performance on ImageNet-1k, ImageNet Real, and ImageNet V2, without any fine-tuning. During the pre-training stage, we adhere to the same experimental settings as described in the original paper.
Furthermore, we juxtapose our method with SVD~\cite{henry19928}, a traditional tensor decomposition technique viable for over-parameterizing student models. Concretely, we employ SVD to substitute MPO within our framework and execute over-parameterization across all parameter matrices of the student model during knowledge distillation. Appendix~\ref{experimental_details} shows more experimental details.

\subsection{Main Experimental Results}
% In this part, we report and analyze the experimental results on NLP and CV tasks.
\paragraph{NLP Tasks}

\begin{table*}[t]
\large
    \centering
    % \caption{Performance comparsion on GLUE benchmark (in percent). "+SVD" represents the use of SVD as the over-parameterization method. "+OPDF" signify the use of OPDF as the over-parameterization method. We report the performance of the model that achieves the best results on the validation set when applied to the test set. Numbers marked with * indicate that their original paper or released code did not test this task, and the result here were reproduced according to the published code. "Corr." means the average of Pearson and Spearman correlation. The best result for each task is highlight in bold. "#Train Params" and "#Inference Params" respectively denote the number of total parameters during training and inference. }
    \caption{Comparison of performance on the GLUE benchmark (in percent). The terms "+SVD" and "+OPDF" represent the use of different over-parameterization methods in a KD model. "\# Train Params" and "\# Inference Params"  refer to the total number of parameters during training and inference, respectively.  Numbers marked with * indicate tasks not tested in the original studies; results here are reproduced from the published code. The best result for each task is highlight in bold.
    For all the results, we report the mean values of five runs using different random seeds.}
        \resizebox{\columnwidth}{!}{
            \begin{tabular}{l|cccccccc|c|cc}
                \hline
                Datasets & \thead{\large RTE \\ \large Acc.} &  \thead{ \large MRPC \\ \large F1/Acc.} &  \thead{\large STS-B \\ \large Corr.} &  \thead{\large CoLA \\ \large Mcc.} &  \thead{\large SST-2 \\ \large F1/Acc.} &  \thead{\large QNLI \\ \large Acc.} & \thead{\large QQP \\ \large F1/Acc.} &  \thead{\large MNLI \\ \large Acc.} & Avg. &  \thead{\large \# Train \\ \large Params \\ \large (M)} & \thead{ \large \# Inference \\ \large Params \\ \large  (M)}\\
                \hline
                BERT-base~\cite{devlin2018bert} & 70.5 & 86.5/81.8 & 86.6 & 54.2 & 92.0 & 91.2 & 88.0/91.0 & 84.2 & 83.4 & 110 & 110 \\
                \hline
                \rowcolor{gray!10}
                \multicolumn{12}{c}{\textbf{BERT-of-Theseus~\cite{xu2020bert}}} \\
                \hline
                None & 65.5 & 85.3/79.6 & 86.2 & 39.2* & 90.4 & 88.7 & 86.1/89.6 & \textbf{81.5} & 79.2 & 66 & 66 \\
                +SVD &  65.5 & 85.4/80.0 & 86.5 & 43.1 & 90.6 & 88.6 & 86.2/89.7 & 80.3 & 79.6 & 90 & 66\\
                +OPDF (Ours)  & \textbf{66.2} & \textbf{85.9/80.5} & \textbf{88.6} & \textbf{45.2} & \textbf{ 91.3} & \textbf{89.0} & \textbf{86.8/90.2} & 81.4 & \textbf{80.5} & 160 & 66 \\
                \hline
                \rowcolor{gray!10}
                \multicolumn{12}{c}{\textbf{LGTM~\cite{ren2023tailoring}}} \\
                \hline                              
                None & 63.3 & 86.3/80.1 & 82.9* & 33.9* & 91.1 & \textbf{89.3} & \textbf{88.0/91.1} & \textbf{82.2} & 78.8 & 67 & 67 \\
                +SVD &  64.7 & 86.8/81.9 & 83.1 & 37.4 & 91.2 & 88.6 & 86.5/89.4 & 79.3 & 78.9 & 91 & 67\\
                +OPDF (Ours) & \textbf{66.9} & \textbf{87.8/82.4} & \textbf{83.3} & \textbf{38.9} & \textbf{91.5} & 88.7 & 87.0/90.2 & 80.9 & \textbf{79.8} & 163 & 67  \\
                \hline
                \rowcolor{gray!10}
                \multicolumn{12}{c}{\textbf{DBKD~\cite{zhou2023bridging}}} \\
                \hline
                None & 61.2 & 83.3/75.5 & / & 25.2 & 88.1 & 86.1 & 85.3/88.7 & 76.1 & 74.4 & 53 & 53 \\
                +SVD &  64.7 & 86.5/78.6 & / & 26.4 & 88.8 & 85.8 & 85.5/89.0 & 76.5 & 75.8 & 69 & 53\\
                +OPDF (Ours) & \textbf{69.1} & \textbf{88.4/83.3} & / & \textbf{27.2} & \textbf{89.8} & \textbf{86.5} & \textbf{86.9/90.2} & \textbf{77.7} & \textbf{77.6} & 83 & 53 \\
                \hline
                \rowcolor{gray!10}
                \multicolumn{12}{c}{\textbf{AD-KD~\cite{wu2023ad}}} \\
                \hline
                None & 68.8 & 88.7/84.3 & \textbf{89.3} & 53.1 & \textbf{91.5} & 90.8 & 85.9/89.5 & 81.7 & 82.4 & 67 & 67 \\
                +SVD &  69.4 & 89.3/85.8 & 88.8 & 53.5 & 89.9 & 90.1 & 86.4/89.8 & 81.5 & 82.6 & 91 & 67\\
                +OPDF (Ours) & \textbf{71.7} & \textbf{90.3/86.8} & 88.9 & \textbf{55.0} & 91.3 & \textbf{91.1} & \textbf{86.8/90.0} & \textbf{82.1} & \textbf{83.4} & 182 & 67 \\
                \hline
                \end{tabular}
        }

    \label{tab:nlp_result}
\end{table*}

We present the results on BERT in Table~\ref{tab:nlp_result}. Firstly, it is evident that integrating KD with over-parameterization methods yields the most significant performance enhancements. Over-parameterization enhances the generalization ability of the student model. Upon comparing the two tensor decomposition techniques, we find that MPO consistently outperforms SVD. This discrepancy arises from the singular value-based SVD in a two-dimensional space, limiting its ability to substantially increase model parameters compared to MPO decomposition~(\eg 90M \emph{vs.} 160M in BERT-of-Theseus). In contrast, MPO allows for arbitrary scaling by increasing the order of decomposition, rendering it more suitable for over-parameterization.
Secondly, following the integration of the OPDF method, the performance of prior KD techniques~(BERT-of-Theseus, LGTM, DBKD, and AD-KD) have exhibited enhancements across a majority of tasks~(\eg RTE, MRPC, CoLA, QQP), while maintaining comparability with the original method in other tasks. This highlights the versatility of OPDF, demonstrating its effectiveness across diverse models and a wide range of tasks. Finally, our findings have revealed that employing the OPDF method can even outperform the performance of the teacher model in MRPC and RTE datasets. This indicates that the process of over-parameterization endows the student model with stronger generalization capabilities, suggesting that employing over-parameterization may offer a potential solution to the bottleneck in current distillation methods where the performance of the student model fails to surpass that of the teacher model.

\paragraph{CV Tasks}
All CV results of our proposed method are shown in Table~\ref{tab:cv_result}. We apply OPDF on three kinds of TinyVit with different total parameters. It is clear that with OPDF, the performance of TinyVit can be significantly improved. In particular, in all datasets, TinyVit applied OPDF is better than vanilla TinyVit. Moreover, TinyVit utilized OPDF with 11M parameters can achieve better performance than TinyVit with 21M parameters. It demonstrates that OPDF is an orthogonal method for various KD methods based on the Transformer whether in the CV or NLP field. Note that since we only involved the over-parameterization procedure in the training phase, the total parameter of the student model will not change in the inference phase. This merit makes the OPDF unique from the existing KD method: one would not increase inference time while enhancing model accuracy and enabling the model to acquire more knowledge from the teacher model.
Moreover, we can observe that the performance of the original TinyVit, SVD over-parameterization, and OPDF over-parameterization improves as the number of parameters gradually increases. This indicates that compared to SVD, the MPO decomposition, which can decompose the parameter matrix to any size, can better enhance the expressive capacity of the student model. The impact of the over-parameterization scale on distillation effectiveness will be analyzed in detail in Section~\ref{scale}.
\begin{table*}[t!]
\footnotesize
    \centering
    % \caption{The linear probe performance on imagenet-1k~\cite{deng2009imagenet}, imagenet Real~\cite{beyer2020we}, imagenet v2~\cite{recht2019imagenet} of TinyVit\cite{wu2022tinyvit} pretrained on imagenet-21k(in percent). Numbers marked with * indicate that these results are got by official checkpoint and released code. The best result for each task is highlight in bold. "\#Train Params" and "\#Inference Params" respectively denote the number of total parameters during training and inference. }
    \caption{The linear probe performance (in percentage) of TinyViT, pre-trained on ImageNet-21k, ImageNet-1k~\cite{deng2009imagenet}, ImageNet Real~\cite{beyer2020we}, and ImageNet v2~\cite{recht2019imagenet}. Numbers marked with * indicate that these results are got by official checkpoint and released code.For all the results, we report the mean values of five runs using different random seeds.}
        \resizebox{\columnwidth}{!}{
            \begin{tabular}{l|cc|cc|cc|cc}
                \hline
                \multirow{2}{*}{\centering Datasets} & \multicolumn{2}{c|}{Imagenet-1k} &  \multicolumn{2}{c|}{Imagenet Real} &  \multicolumn{2}{c|}{Imagenet V2} &  \thead{\# Train Params} &  \thead{\# Inference Params} \\
                  &  top-1 &  top-5 &  top-1 &  top-5 & top-1 & top-5 & (M) & (M)  \\
                \hline
                CLIP-ViT-L/14~\cite{radford2021learning} & 84.8* & / & 88.9* & / & 75.1* & / & 321 & 321 \\
                \hline
                \rowcolor{gray!10}
                \multicolumn{9}{c}{\textbf{TinyVit-5M~\cite{wu2022tinyvit}}} \\
                \hline
                None & 77.4* & 94.1* & 86.1* & 97.5* & 66.8* & 87.6* & 5.4 & 5.4 \\
                +SVD & 77.9 & 95.1 & 86.3 & 97.3 & 68.7 & 88.4 & 7.6 & 5.4 \\
                +OPDF (Ours) & \textbf{80.0} & \textbf{96.7} & \textbf{87.4} & \textbf{98.1} & \textbf{ 69.4} & \textbf{88.9} & 9.9 & 5.4 \\
                \hline
                \rowcolor{gray!10}
                \multicolumn{9}{c}{\textbf{TinyVit-11M}} \\
                \hline
                None & 80.5* & 95.6* & 87.8* & 98.0* & 70.7* & 90.4* & 11 & 11 \\
                +SVD &  82.0 & 96.7 & 88.4 & 97.9 & 71.7 & 91.4 & 17 & 11 \\
                +OPDF (Ours) & \textbf{82.5} & \textbf{96.9} & \textbf{88.9} & \textbf{98.3} & \textbf{ 72.4} & \textbf{92.6} & 23 & 11 \\
                \hline
                \rowcolor{gray!10}
                \multicolumn{9}{c}{\textbf{TinyVit-21M}} \\
                \hline
                None & 82.3* & 96.3* & 88.9* & 98.3* & 73.0* & 91.9* & 21 & 21 \\
                +SVD & 82.9 & 96.8 & 88.3 & 97.8 & 71.8 & 92.4 & 29 & 21 \\
                +OPDF (Ours) & \textbf{84.0} & \textbf{97.5} & \textbf{89.4} & \textbf{98.4} & \textbf{74.9} & \textbf{93.4} & 38 & 21 \\
                \hline
                \end{tabular}
        }

    \label{tab:cv_result}
\end{table*}

\subsection{Further Analysis}
\paragraph{Performance Comparison \emph{w.r.t.} Parameter Increasing Rate.}
\label{scale}
Our OPDF method facilitates the flexible expansion of model parameters, thereby highlighting the significance of the parameter increase rate on model performance. Consequently, we investigate the influence of this rate on model efficiency further. To underscore the general applicability of our findings, we intentionally over-parameterize two models: DBKD and LGTM. We then elucidate their relationship with fine-tuning performance on MRPC tasks. All results are depicted in Figure~\ref{Scale_factor}.

It is observed that the performance of both the LGTM~\cite{ren2023tailoring} and DBKD~\cite{zhou2023bridging} models on the MRPC task consistently improves with an increase in parameters. This enhancement substantiates the efficacy of using the OPDF for over-parameterizing models, which in turn significantly boosts the performance of knowledge distillation models. Furthermore, after over-parameterization, the performance of the models is capable of achieving, at a minimum, the level of their original benchmarks~(\eg 83.3 for DBKD and 86.3 for LGTM).
The enhancement of model performance through over-parameterization has its limitations. As demonstrated in Figure~\ref{Scale_factor}, beyond certain thresholds of over-parameterization~(\eg 1.6$\times$ for DBKD and 2.5$\times$ for LGTM), the performance improvements of the models no longer exhibit significant gains. This observation indicates that there are inherent limits to the benefits that can be achieved through over-parameterization in knowledge distillation models. These limits are likely influenced by structural characteristics of each model and size of the initial model configuration.

% \begin{figure}[tb]
\begin{figure}[t!]
\vspace{-12pt}
    \centering
    \hfill
    \subfigure[\small Impact of scale factor]{
    \label{Scale_factor}
    \begin{minipage}[b]{0.31\linewidth}
        \centering
        \includegraphics[width=\linewidth]{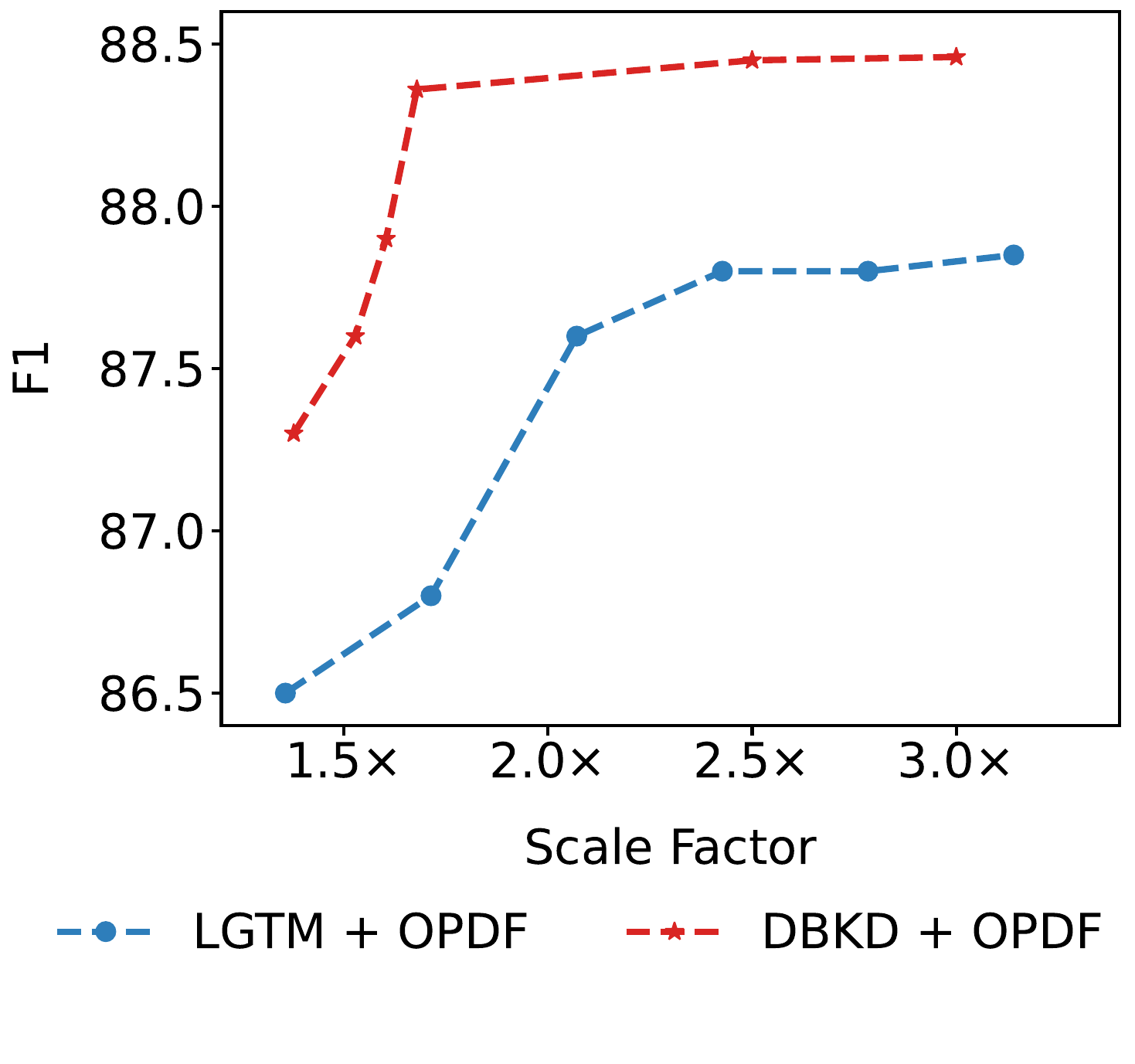}
        \label{Scale_factor}
        \vspace{-6pt}
    \end{minipage}
    }
    \hfill
    \subfigure[\small Impact of learning rate]{
    \label{learning_rate}
    \begin{minipage}[b]{0.31\linewidth}
        \centering
        \includegraphics[width=\linewidth]{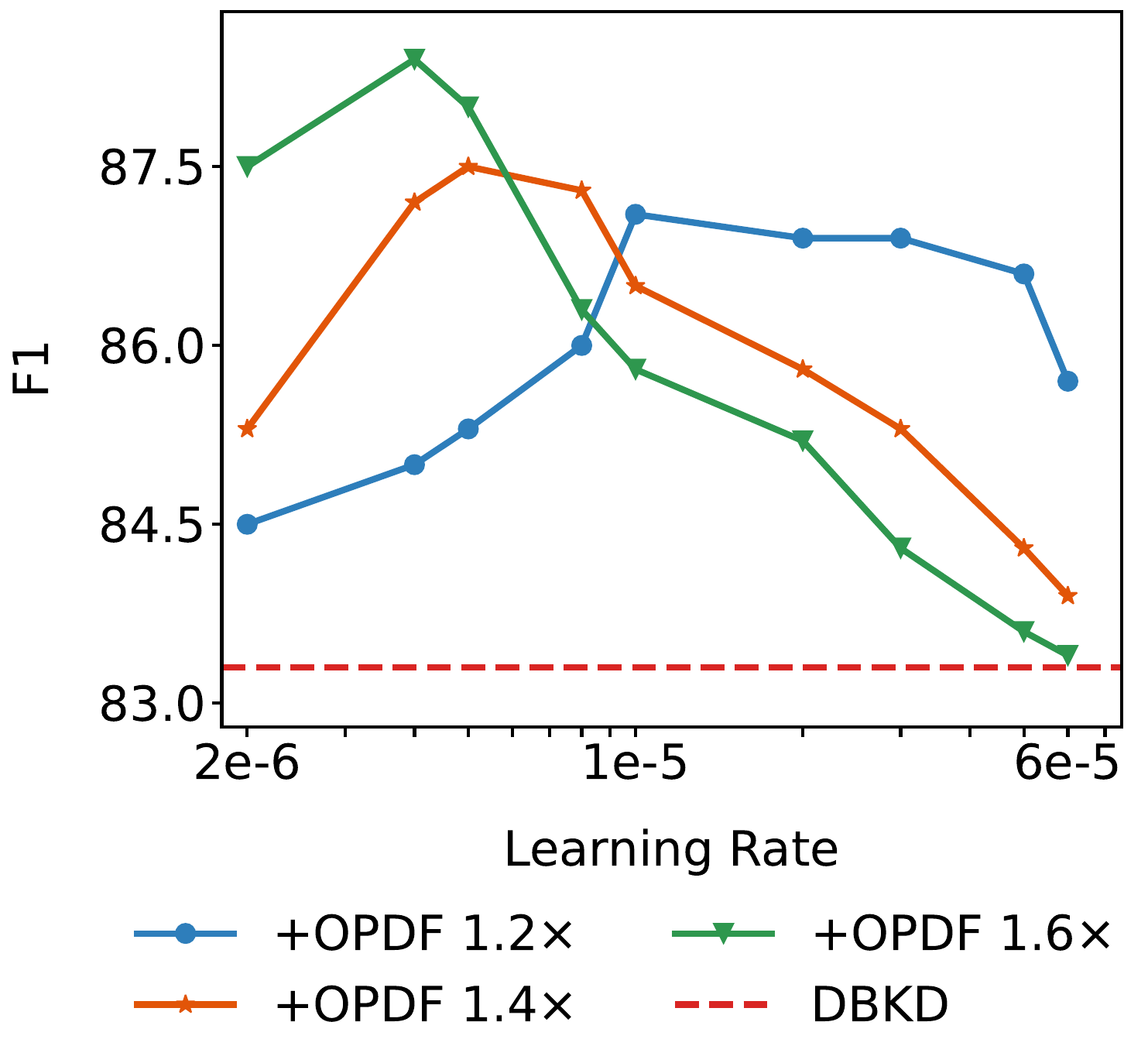}
        \label{learning_rate}
        \vspace{-6pt}
    \end{minipage}
    }
    \hfill
    \subfigure[\small Impact of OPDF components]{
    \label{ablation}
    \begin{minipage}[b]{0.31\linewidth}
        \centering
        \includegraphics[width=\linewidth]{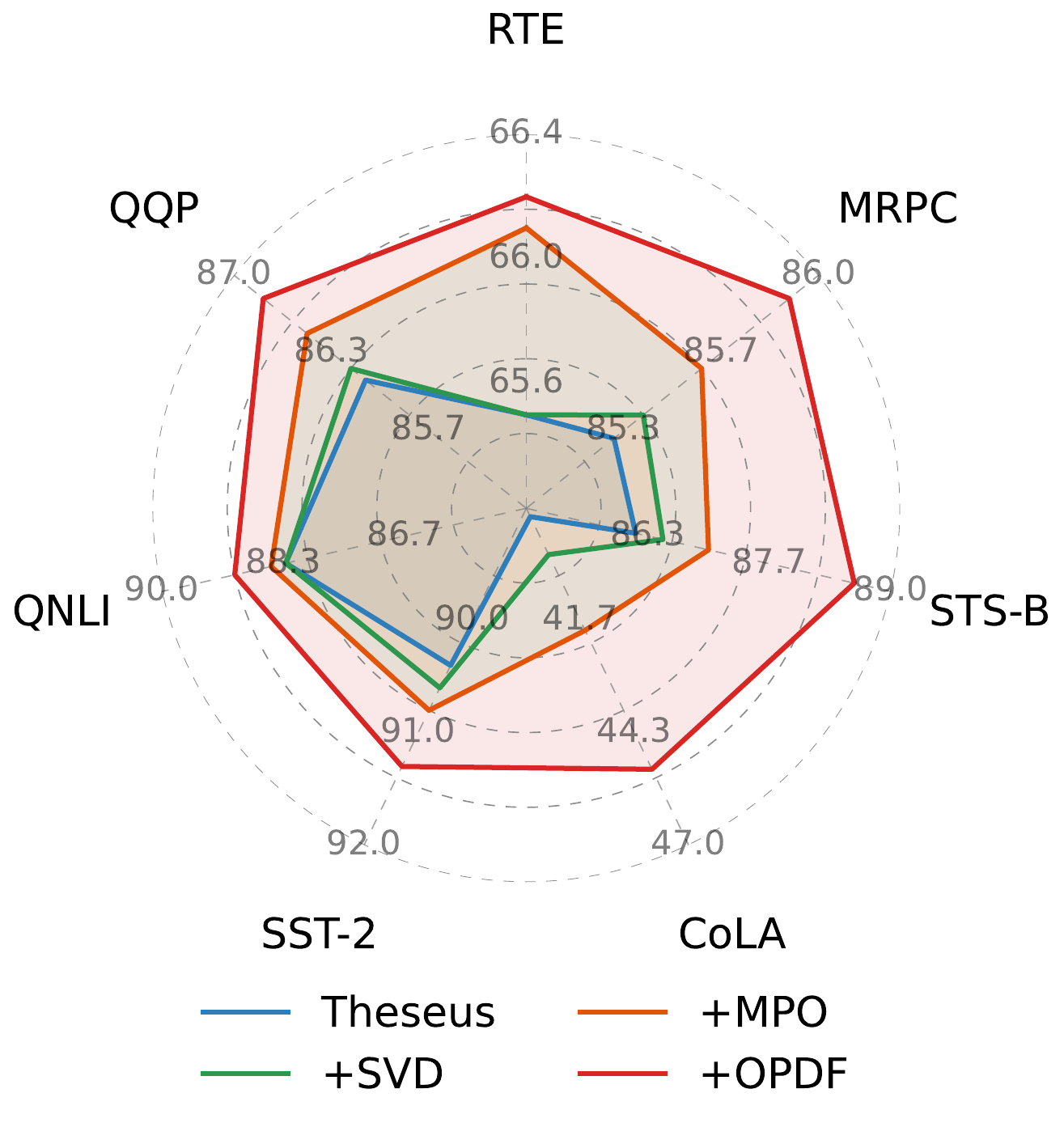}
        \label{ablation}
        \vspace{-6pt}
    \end{minipage}
    }    
    \caption{The impact of over-parameterization scale, learning rate, and various components of the OPDF on distillation model performance is explored. Figure~\ref{Scale_factor} demonstrates the performance of the LGTM and DBKD model on the MRPC task following the implementation of the OPDF. Figure~\ref{learning_rate} presents the performance of DBKD + OPDF with different over-parameterization scales on the MRPC task. Figure~\ref{ablation} displays the performance of the theseus model across various tasks, utilizing different over-parameterization methods and integrating various components of the OPDF.}
    \label{fig-over_parametrization}
    \vspace{-6pt}
\end{figure}

\paragraph{Hyper-parameters Tuning}
\label{Hyper_parameters_Tuning}
OPDF decomposes the original weight tensor through over-parameterization, leading to the updating of more parameters. Consequently, the tensor product results in larger updates to the existing parameters in the backward phase. In Figure~\ref{learning_rate}, we illustrate the relationship between the performance on the MRPC task and learning rate when the parameters of the DBKD model are expanded to 1.2$\times$, 1.4$\times$, and 1.6$\times$ their original size.

There exists an optimal learning rate for every scale of over-parameterization. Deviating from this optimal rate, whether by increasing or decreasing the learning rate, results in diminished model performance. The reduction in performance due to a lower learning rate can be attributed to the model becoming trapped in a local optimum. 

Additionally, we observe that peak model performance consistently increases with the scale of over-parameterization. This finding aligns with the conclusions drawn from Figure~\ref{Scale_factor}. Moreover, as the scale of over-parameterization increases, the learning rate required to achieve optimal model performance decreases. This occurs because using the tensor product to restore the shape of the tensors to that of the original weight tensors also scales the updated values, resulting in significant changes. Consequently, an increasing learning rate leads to declining performance in the KD model, indicating that the learning rate should decrease as the over-parameterization scale increases.

Finally, despite changes in the learning rate, the performance of the model with OPDF consistently remains at least as high as that of the original method. This indicates that OPDF is not sensitive to learning rate variations during the distillation stage. This resilience is due to OPDF's ability to factorize the parameter matrix in almost lossless conditions, ensuring that the decomposed matrix can match or exceed the training effectiveness of the original matrix without introducing errors.

\paragraph{Impact of MPO strategy}

To demonstrate the robustness of our MPO methods, we applied different MPO methods to the DBKD and AD-KD model on the RTE, MRPC, STS-B, CoLA, and SST-2 task. The experimental results are presented in Table~\ref{tab:mpo_method}. To maintain consistent over-parameterization scales, we used the same decomposition scale~(L) for each KD model across the same task.
\begin{table*}[h]
\tiny
    \centering
    \caption{Comparison of performance on the GLUE benchmark (in percent). In the tensor representation, "L" denotes the number of "1"s in the dimension list.}
    \resizebox{0.6\columnwidth}{!}{
            \begin{tabular}{l|ccccc}
                \hline
                \small Experiments & \thead{\small RTE \\ \small Acc.} &  \thead{\small MRPC \\ \small F1/Acc.} &  \thead{\small STS-B \\ \small Corr.} &  \thead{\small CoLA \\ \small Mcc.} &  \thead{\small SST-2 \\ \small F1/Acc.} \\
                \hline
                \rowcolor{gray!10}
                \multicolumn{6}{c}{ \small \textbf{DBKD}} \\
                \hline
                \small $L$ & \small 4 &  \small 8 &  \small / &  \small 7 &  \small 4 \\
                $\Tensor{T}_{64,L,48}^{32,L,24}$ & \small 69.1 & \small 88.4/83.3 & \small / & \small 27.2 & \small 89.8 \\
                $\Tensor{T}_{32,2,L,2,24}^{16,2,L,2,12}$ & \small 68.0 & \small 86.3/81.0 & \small / & \small 25.2 & \small 89.0 \\
                $\Tensor{T}_{8,4,2,L,2,4,6}^{4,4,2,L,2,3,4}$ & \small 68.5 & \small 87.9/82.5 & \small / & \small 26.1 & \small 88.9 \\
                \hline
                \rowcolor{gray!10}
                \multicolumn{6}{c}{ \small \textbf{AD-KD}} \\
                \hline
                \small $L$ & \small 8 & \small 3 & \small 6 & \small 1 & \small 3 \\
                $\Tensor{T}_{64,L,48}^{32,L,24}$ & \small 71.7 & \small 90.3/86.8 & \small 88.9 & \small 55.0 & \small 91.3 \\
                $\Tensor{T}_{32,2,L,2,24}^{16,2,L,2,12}$ & \small 70.9 & \small 89.6/86.1 & \small 88.7 & \small 54.4 & \small 89.2 \\
                $\Tensor{T}_{8,4,2,L,2,4,6}^{4,4,2,L,2,3,4}$ & \small 71.0 & \small 89.8/86.4 & \small 88.3 & \small 54.9 & \small 90.4 \\
                \hline
                \end{tabular}
        }

    \label{tab:mpo_method}
\end{table*}

We can observe that the performance of our approach consistently stabilizes around certain values, indicating that our method is not sensitive to the specific MPO techniques used. Therefore, when over-parameterizing, we should focus primarily on the decomposition scale rather than the MPO method employed.

\vspace{-3pt}
% 做消融实验
\paragraph{Ablation Study}
Our approach consists of two novel improvements: (1) the over-parameterization procedure for the student model, (2) the distillation loss for auxiliary tensors for effective training. To verify the effectiveness of each component, we conduct the ablation study on the GLUE benchmark to analyze the contribution of each part. We consider removing over-parameterization and distillation loss respectively. The ablation results of our OPDF are shown in Figure~\ref{ablation}. 

Firstly, it is clear that regardless of the over-parameterization method used, the area of the radar chart is greater than that of the vanilla theseus. This outcome suggests that over-parameterization can greatly improve the performance of distillation methods. Secondly, further analysis of the different over-parameterization methods reveals that MPO consistently outperforms SVD across all datasets. This improvement is attributed to MPO's ability to decompose parameter matrices into higher orders, effectively enlarging the size of the parameter matrix. Lastly, we examine the contribution of the $L_{Aux}$ term. The radar chart area is significantly larger when OPDF is utilized in conjunction with $L_{Aux}$ than with MPO alone. This indicates that $L_{Aux}$ effectively enhances knowledge transfer from the teacher model. The underlying reason for this phenomenon is that over-parameterized models can concentrate on learning central tensors containing critical information, while the $L_{Aux}$ term assists in aligning auxiliary tensors.
We can see that removing any component would lead to a decrease in the model performance. It shows the effectiveness of all these components in our approach.

\vspace{-3pt}
\section{Conclusion}
In this paper, we proposed OPDF, a novel over-parameterization distillation framework designed to enhance the effectiveness of knowledge distillation. This framework employs MPO as a tensor decomposition technique to expand small models into larger ones, thereby bridging the capacity gap between the teacher and student models.
Moreover, to enhance the effectiveness of knowledge distillation, our proposed OPDF framework introduces a tensor constraint loss. The OPDF framework utilizes MPO to decompose each weight matrix into a central tensor and auxiliary tensors. By aligning the auxiliary tensors, OPDF not only facilitates the transfer of crucial knowledge from the teacher model but also preserves the student model's ability to think independently. This approach provides the student model with the potential to outperform the teacher model.
Our ablation studies demonstrated that all components of the OPDF contribute to enhancing the effectiveness of knowledge distillation. Experimental results across various tasks in natural language processing and computer vision domains validate the efficacy of our proposed method in improving model distillation. 
Although the number of parameters was increased by MPO during training, the factorized matrices can be merged to reorganize the original parameter matrix in almost lossless conditions. This means that OPDF can enhance the performance of the distillation model without increasing the inference latency. Moreover, since OPDF is based on tensor decomposition, it is orthogonal to most distillation methods.

In our future work, we will investigate more efficient and effective tensor decomposition methods for student model over-parameterization. In addition, we will also apply OPDF to other important backbone models, such as in the multimodal learning domains.

\section*{Impact statement}
\label{impacts}
This paper proposes a novel knowledge distillation framework for model compression field, which is helpful to reduce storage requirements and computational complexity. This method facilitates the practical deployment of models in real-world applications and supports energy conservation. We focus exclusively on over-parameterizing small student models, presenting no potential ethical risks.

%\medskip
\section*{Acknowledgement}
This work was supported by the National Natural Science Foundation of China (No. 62476278, No. 62206299, No. 92270118, and No. 11934020) and the Beijing Natural Science Foundation (No. 1232009). H.S would like to acknowledge the support from the Fundamental Research Funds for the Central Universities (No. 202230265).

\bibliographystyle{unsrt}
\bibliography{neurips_2024}

%%%%%%%%%%%%%%%%%%%%%%%%%%%%%%%%%%%%%%%%%%%%%%%%%%%%%%%%%%%%

% \bibliographystyle{unsrt}
% \bibliography{neurips_2024}
% \appendix
% \newpage
\clearpage
\newpage
\appendix
\include{appendix}

{\centering \Large \textbf{APPENDIX}}
\vspace{6pt}

\setcounter{figure}{0}
\setcounter{table}{0}
\setcounter{equation}{0}
\setcounter{algorithm}{0}

\renewcommand{\theequation}{S.\arabic{equation}}
\renewcommand{\thefigure}{S.\arabic{figure}}
\renewcommand{\thetable}{S.\arabic{table}}
\renewcommand{\thealgorithm}{S.\arabic{algorithm}}

\section{Tensor and Matrix Product Operators}
\label{appendix_a}
As introduced in~\cite{gao2020compressing}, the concept of a tensor is specified as:

\paratitle{Definition1} \\
(Tensor). Let $D_1, D_2..., D_N \in N$ denote index upper bounds. A tensor $\Tensor{T}\in \mathbb{R}^{D_1,...,D_n}$ of order $N$ is an $N$-way array where elements $\Tensor{T}_{d_1,d_2,...,d_n}$ are indexed by $d_n\in\{1,2,...,D_n\}$ for $1\leq n\leq N$ 

\paratitle{Definition2} \\
(Matrix product operator).
We can reshape a matrix to high order tensor, denoted as:
\begin{equation}
    \Matrix{M}_{x\times y} = \Matrix{M}_{i_1i_2...i_n,j_1j_2...j_n}
\end{equation}
Here, the one-dimensional coordinate $x$ of the input signal $\mathbf{x}$ with dimension $N_x$ is reshaped into a coordinate in a $n$-dimensional space, labeled by $(i_1 i_2 \cdots i_n)$.
Hence, there is a one-to-one mapping between $x$ and $(i_1 i_2 \cdots i_n)$.
Similarly, the one-dimensional coordinate $y$ of the output signal $\mathbf{y}$ with dimension $N_y$ is also reshaped into a coordinate in a $n$-dimensional space, and there is a one-to-one correspondence between $y$ and $(j_1j_2\cdots j_n)$.
If $I_k$ and $J_k$ are the dimensions of $i_k$ and $j_k$, respectively, then
\begin{equation}
    \prod_{k=1}^{n}I_{k} = N_{x}, \quad \prod_{k=1}^{n}J_{k} = N_{y}  . \label{eq:dims}
\end{equation}

The MPO representation of $M$ is obtained by factorizing it into a product of $n$ local tensors
\begin{eqnarray}
 M_{i_1\cdots i_n,j_1\cdots j_n} =  \Tensor{T}^{(1)} [i_1,j_1] \cdots \Tensor{T}^{(n)} [i_n,j_n]  \label{Eq:MPO}
\end{eqnarray}
where $\Tensor{T}^{(k)}[j_k,i_k]$ is a $D_{k-1}\times D_{k}$ matrix with $D_k$ the virtual basis dimension on the bond linking $\Tensor{T}^{(k)}$ and $\Tensor{T}^{(k+1)}$ with $D_0=D_n=1$.

\section{Theorem}
\label{appendix_b}
\begin{thm}\label{thm1}
Suppose that the tensor $\textbf{W}^{(k)}$ of matrix $W$ that is satisfy 
\begin{align}
    &\Matrix{W} = \Matrix{W}^{(k)} + \Matrix{E}^{(k)}  ,D( \textbf{W}^{(k)}) = d_k,\notag\\
    &where\quad||\Matrix{E}^{(k)}||_F^2 = \epsilon_k^2 , k = 1,...,d-1. 
\end{align}
Then $MPO~(\Matrix{W})$ with the $k$-th bond dimension $d_k$ upper bound of truncation error satisfy:
\begin{equation}
    ||\Matrix{W}-MPO~(\Matrix{W})||_F \leq \sqrt{\sum_{k=1}^{d-1} \epsilon_k^2}
\end{equation}
\end{thm}

$Proof.$ The proof is by induction. For $n = 2$ the statement follows from the properties of the SVD. Consider an arbitrary $n > 2$. Then the first unfolding $\Matrix{W}^{(1)}$ is decomposed as
\begin{equation}
    \Matrix{W}^{(1)} = \Matrix{U}_1 \Matrix{\lambda}_1 \Matrix{V}_1 + \Matrix{E}^{(1)} = \Matrix{U}_1\Matrix{B}^{(1)} + \Matrix{E}^{(1)}
\end{equation}
where $\Matrix{U}_1$ is of size $r_1\times i_1 \times j_1$ and $|| \Matrix{E}^{(1)}||_F^2 = \epsilon_1^2$. The matrix $\Matrix{B}_1$ is naturally associated with a $(n-1)$-dimensional tensor $\Tensor{B}^{(1)}$ with elements $\Tensor{B}^{(1)}(\alpha , i_2,j_2, ..., i_n,j_n)$, which will be decomposed further. This means that $\Matrix{B}_1$ will be approximated by some other matrix $\hat{\Matrix{B}_1}$. From the properties of the SVD it follows that $\Matrix{U}_1^{T}\Matrix{E}^{(1)}=0$, and thus

\begin{align}
    & ||\Matrix{W}-\Tensor{B}^{(1)}||^2_F \notag\\ 
    & = ||\Matrix{W}_1 - \Matrix{U}_1\hat{\Matrix{B}_1}||_F^{2} \notag\\
    & = ||\Matrix{W}_1 - \Matrix{U}_1(\hat{\Matrix{B}_1} + \Matrix{B}_1 - \Matrix{B}_1)||_F^{2}\notag\\
    & = ||\Matrix{W}_1 - \Matrix{U}_1\Matrix{B}_1||_F^{2} + ||\Matrix{U}_1(\hat{\Matrix{B}_1} - \Matrix{B}_1)||_F^2
\end{align}

and since $\Matrix{U}_1$ has orthonormal columns,
\begin{equation}
    ||\Matrix{W}-\Tensor{B}^{(1)}||_F^2 \leq \epsilon_1^2 + ||\Matrix{B}_1-\hat{\Matrix{B}_1}||_F^2.
    \label{eq:thm1}
\end{equation}
and thus it is not difficult to see from the orthonormality of columns of $\Matrix{U}_1$ that the distance of the $k$-th unfolding $(k=2,...,d_k-1)$ of the $(d-1)$-dimensional tensor $\Tensor{B}^{(1)}$ to the $d_k$-th rank matrix cannot be larger than $\epsilon_k$. Proceeding by induction, we have 
\begin{equation}
    ||\Matrix{B}_1 - \hat{\Matrix{B}_1} ||_F^2 \leq \sum_{k=2}^{d-1} \epsilon_k^2,
\end{equation}
combine with Eq.~\eqref{eq:thm1}, this complets the proof.

\section{Algorithms}

The over-parameterized distillation framework algorithm is shown in Algorithm~\ref{alg:over-all-alg}.
\floatname{algorithm}{Algorithm}
\begin{algorithm}
\small
    \caption{Over-parameterized distillation framework.}   
    \begin{algorithmic}[1] %每行显示行号
        \Require The parameter matrices list of student model  $\Matrix{\{M_s(k)\}_{k=1}^{n}}$, the parameter  matrices list of teacher model $\Matrix{\{M_t(k)\}_{k=1}^{m}}$.
        \For{$k=1 \to n$}
            \State Select $\Matrix{M_t(k_t)}$ which has same shape as $\Matrix{M_s(k)}$.
            \State $\Matrix{M_s(k)} \to \rm{MPO}~(\Matrix{M_s(k)})$.            
            \State $\Matrix{M_t(k_t)} \to \rm{MPO}~(\Matrix{M_t(k_t)})$.
        \EndFor
        \Repeat
          \State Compute $L_{Aux}$ between $\{\rm{MPO}~(\Matrix{M_s(k)})\}_{k=1}^{n}$ and $\{\rm{MPO}~(\Matrix{M_t(k_t)})\}_{k_t=1}^{m_t}$ by using Eq.~\eqref{eq:loss_auxiliary}.
          \State Compute distill loss $L_{distill}$.
          \State Backward $L_{Aux}$ and $L_{distill}$.
        \Until{Student model converges}
    \end{algorithmic}
\label{alg:over-all-alg}
\end{algorithm}

The MPO pseudocode is shown in Algorithm~\ref{alg:mpo-decomposition}.

\floatname{algorithm}{Algorithm}
\begin{algorithm}
\small
    \caption{MPO decomposition for a matrix.}   
    \begin{algorithmic}[1] %每行显示行号
        \Require matrix $\Matrix{M}$, the number of local tensors $n$
        \Ensure: MPO tensor list $ \{\Tensor{T}_{(k)}\}_{k=1}^{n}$
        \For{$k=1 \to n-1$ }
            \State $\Matrix{M}[I,J]\longrightarrow \Matrix{M}[d_{k-1}\times i_k\times j_k,-1]$
            \State $\Matrix{U}\lambda \Matrix{V}^\top=\rm{SVD}~(\Matrix{M})$
            \State $\Matrix{U}[d_{k-1}\times i_k\times j_k, d_k]\longrightarrow \Tensor{U}[d_{k-1},i_k,j_k,d_k]$
            \State $\Tensor{T}^{(k)}:= \Tensor{U} $
            \State $\Matrix{M}:=\lambda \Matrix{V}^{\top}$
        \EndFor
        \State $\Tensor{T}^{(n)}:=\Matrix{M}$
        \State $\rm{Normalization}$
        \State \Return{$ \{\Tensor{T}_{(k)}\}_{k=1}^{n}$}
    \end{algorithmic}
\label{alg:mpo-decomposition}
\end{algorithm}

\section{Addition Experiment Results}

\subsection{Memory and time cost}
\label{cost}

The distillation cost (memory and time cost) of the original model and the model after applying OPDF are shown in Table~\ref{tab:gpu_time}. We can observe that as the number of parameters obtained from MPO decomposition increases, both the training time and memory cost increase. However, as the dataset size increases, the ratio of additional time and memory required for training by OPDF to the original training requirements generally exhibits a decreasing trend (e.g., 0.6/0.4 for RTE vs 0.3/0.1 for MNLI in BERT-of-Theseus model). Therefore, the additional time and memory introduced by our method become less of a critical bottleneck affecting the training speed as the dataset size increases.
\begin{table*}[h]
\large
    \centering
    \caption{Training time and Memory Cost. (Train time(S) / Memory Cost(GB))}
        \resizebox{\columnwidth}{!}{
            \begin{tabular}{l|cccccccc}
                \hline
                Datasets & \thead{\large RTE} &  \thead{ \large MRPC} &  \thead{\large STS-B} &  \thead{\large CoLA} &  \thead{\large SST-2} &  \thead{\large QNLI } & \thead{\large Q QP } &  \thead{\large MNLI }\\
                \hline
                \rowcolor{gray!10}
                \multicolumn{9}{c}{\textbf{BERT-of-Theseus}} \\
                \hline
                None & 400.2/14.8 & 739.5/14.8 & 754.4/14.8 & 1365.4/14.8 & 2553.7/14.1 & 3514.2/14.1 & 7518.7/14.1 & 8873.9/13.5 \\
                +OPDF (Ours) & 625.8/20.7 & 1054.9/20.6 & 1932.0/20.5 & 2560.4/16.8 & 21864.5/22.1 & 5041.6/27.0 & 10301.9/18.9 & 11674.8/14.8 \\
                \hline
                \rowcolor{gray!10}
                \multicolumn{9}{c}{\textbf{LGTM}} \\
                \hline                              
                None & 1086.1/9.7 & 1408.5/9.6 & 2049.1/9.6 & 3358.9/9.6 & 5270.9/9.6 & 8142.8/9.6 & 30272.6/9.6 & 31554.8/9.6 \\
                +OPDF (Ours) &  1976.3/19.7 & 2611.9/19.6 & 3603.6/19.7 & 2348.4/19.6 & 9983.4/19.6 & 14838.4/19.6 & 44058.5/14.7 & 47849.4/19.6 \\
                \hline
                \rowcolor{gray!10}
                \multicolumn{9}{c}{\textbf{DBKD}} \\
                \hline
                None & 40.7/2.1 & 80.3/3.2 & NA & 186.4/3.2 & 1355.8/3.2 & 2149.2/3.2 & 7487.5/3.2 & 15513.6/3.2 \\
                +OPDF (Ours) & 93.1/5.0 & 213.0/6.6 & NA & 373.3/6.2 & 2793.8/5.4 & 6076.5/6.6 & 14030.7/6.6 & 21273.4/5.0 \\
                \hline
                \rowcolor{gray!10}
                \multicolumn{9}{c}{\textbf{AD-KD}} \\
                \hline
                None & 308.5/3.8 & 351.3/3.8 & 495.6/3.8 & 780.3/5.9 & 3637.4/5.9 & 5832.7/5.9 & 28763.3/5.9 & 41898.4/20.6 \\

                +OPDF (Ours) & 1156.8/14.4 & 1391.3/14.5 & 1604.2/12.1 & 2249.8/18.1 & 8802.3/14.5 & 13551.1/14.5 & 63695.5/14.5 & 65735.8/34.0 \\
                \hline
                \end{tabular}
                }

    \label{tab:gpu_time}
\end{table*}

We show the time of overparameterization using MPO and the contraction of decomposed matrices into the original matrix in Table~\ref{tab:mpo_time} as follows.  It can be observed that the time required for decomposition and reconstruction is acceptable compared to the training duration.
\begin{table*}[h]
\large
    \centering
    \caption{The spending time (s) of decomposing and reconstructing.}
        \resizebox{0.7\columnwidth}{!}{
            \begin{tabular}{l|cccccccc}
                \hline
                Datasets & \thead{\large RTE} &  \thead{ \large MRPC} &  \thead{\large STS-B} &  \thead{\large CoLA} &  \thead{\large SST-2} &  \thead{\large QNLI } & \thead{\large QQP } &  \thead{\large MNLI }\\
                \hline
                \rowcolor{gray!10}
                \multicolumn{9}{c}{\textbf{BERT-of-Theseus}} \\
                \hline
                Decompose & 397.4 & 308.5 & 400.2 & 154.6 & 797.7 & 671.6 & 584.7 & 137.5 \\
                Reconstruct & 2.3 & 2.0 & 2.4 & 0.7 & 12.8 & 3.3 & 10.9 & 0.8 \\
                \hline
                \rowcolor{gray!10}
                \multicolumn{9}{c}{\textbf{LGTM}} \\
                \hline                              
                Decompose & 403.6 & 369.5 & 377.8 & 192.2 & 131.4 & 123.9 & 80.2 & 83.0 \\
                Reconstruct & 8.6 & 6.8 & 7.2 & 2.6 & 1.0 & 0.9 & 1.0 & 1.5 \\
                \hline
                \rowcolor{gray!10}
                \multicolumn{9}{c}{\textbf{DBKD}} \\
                \hline
                Decompose & 117.5 & 189.1 & NA & 168.5 & 153.1 & 166.2 & 110.7 & 165.0 \\
                Reconstruct & 0.9 & 1.0 & NA & 0.8 & 0.9 & 0.8 & 0.7 & 0.8 \\
                \hline
                \rowcolor{gray!10}
                \multicolumn{9}{c}{\textbf{AD-KD}} \\
                \hline
                Decompose & 291.7 & 313.0 & 232.6 & 235.5 & 119.0 & 148.0 & 148.1 & 171.6 \\
                Reconstruct & 1.6 & 1.8 & 1.2 & 1.4 & 0.9 & 1.0 & 2.5 & 1.2 \\
                \hline
                \end{tabular}
        }

    \label{tab:mpo_time}
\end{table*}

\subsection{Experimental Details}
\label{experimental_details}

%先介绍怎么表示张量.然后说张量对齐.再说nlp张量分解结果展示在table 1.最后提一嘴cv的实验中超参数展示在表2.
As illustrated in Eq.~\eqref{eq:mpo}, when a parameter matrix $\Matrix{W}$ is given, its MPO decomposition into a product of $n$ local tensors can be represented as follows:
\begin{equation}
    \textsc{MPO}~(\Matrix{W})= \Tensor{T}_{j_1,j_2,j_3,...,j_n}^{i_1,i_2,i_3,...,i_n}.
    \label{eq:mpo2}
\end{equation}
% Bert of theseus~\cite{xu2020bert}, LGTM~\cite{ren2023tailoring}, DBKD~\cite{zhou2023bridging} and AD-KD~\cite{wu2023ad} are based on Bert which is constructed by transformer block. We decompose the feed-forward network and multi-head attention layer in the transformer block. Moreover, the teacher model should be applied the same decomposition size as the student model for auxiliary tensors alignment. When computing $\mathcal{L}_{Aux}$, we align the n-th layer of the student model with the N-th layer of the teacher model, where N is generally an integer multiplied by n. The detailed hyper-parameter setting of the NLP distillation model was shown in Table\ref{tab:hyper_nlp_att} and Table\ref{tab:hyper_nlp_ffn}.

The models mentioned—Bert of theseus~\cite{xu2020bert}, LGTM~\cite{ren2023tailoring}, DBKD~\cite{zhou2023bridging} and AD-KD~\cite{wu2023ad}—are all variants of BERT, which itself is built using transformer blocks.  We decompose both the feed-forward network and the multi-head attention layer within the transformer block. Moreover, the teacher model must employ a decomposition granularity that is consistent with that of the student model to ensure proper alignment of auxiliary tensors. When calculating the auxiliary loss $\mathcal{L}_{Aux}$, the alignment is typically between the n-th layer of the student model and the N-th layer of the teacher model, where N is generally an integer multiple of n. The detailed hyperparameter settings for these NLP distillation models are provided in Table~\ref{tab:hyper_nlp_ffn}and~\ref{tab:hyper_nlp_att}. In NLP tasks, our method takes half to two GPU hours on A100 GPU.
\begin{table*}[h]
    \centering
    \caption{The feed-forward network layer settings in NLP distilation model.}
        \resizebox{\columnwidth}{!}{
            \begin{tabular}{l|cccc}
                \hline
                Experiments & \thead{ RTE} &  \thead{ MRPC} &  \thead{ STS-B} &  \thead{ CoLA} \\
                \hline
                \rowcolor{gray!10}
                \multicolumn{5}{c}{\textbf{BERT-of-Theseus~\cite{xu2020bert}}} \\
                \hline
                SVD & $\Tensor{T}_{64,48}^{32,24}$ & $\Tensor{T}_{64,48}^{32,24}$ & $\Tensor{T}_{64,48}^{32,24}$ & $\Tensor{T}_{64,48}^{32,24}$\\
                OPDF (Ours) & $\Tensor{T}_{64,1,1,1,48}^{32,1,1,1,24}$ & $\Tensor{T}_{64,1,1,1,48}^{32,1,1,1,24}$ & $\Tensor{T}_{64,1,1,1,48}^{32,1,1,1,24}$ & $\Tensor{T}_{64,1,48}^{32,1,24}$\\
                \hline
                \rowcolor{gray!10}
                \multicolumn{5}{c}{\textbf{LGTM~\cite{ren2023tailoring}}} \\
                \hline
                SVD & $\Tensor{T}_{64,48}^{32,24}$ & $\Tensor{T}_{64,48}^{32,24}$ & $\Tensor{T}_{64,48}^{32,24}$ & $\Tensor{T}_{64,48}^{32,24}$\\
                OPDF (Ours) & $\Tensor{T}_{64,1,1,1,48}^{32,1,1,1,24}$ & $\Tensor{T}_{64,1,1,1,48}^{32,1,1,1,24}$ & $\Tensor{T}_{64,1,1,1,48}^{32,1,1,1,24}$ & $\Tensor{T}_{64,1,1,1,1,48}^{32,1,1,1,1,24}$\\
                \hline
                \rowcolor{gray!10}
                \multicolumn{5}{c}{\textbf{DBKD~\cite{zhou2023bridging}}} \\
                \hline
                SVD & $\Tensor{T}_{64,48}^{32,24}$ & $\Tensor{T}_{64,48}^{32,24}$ & / & $\Tensor{T}_{64,48}^{32,24}$\\
                OPDF (Ours) & $\Tensor{T}_{64,1,1,1,1,48}^{32,1,1,1,1,24}$ & 
                $\Tensor{T}_{64,1,1,1,1,1,1,1,1,48}^{32,1,1,1,1,1,1,1,1,24}$ & 
                / & 
                $\Tensor{T}_{64,1,1,1,1,1,1,1,48}^{32,1,1,1,1,1,1,1,24}$\\
                \hline
                \rowcolor{gray!10}
                \multicolumn{5}{c}{\textbf{AD-KD~\cite{wu2023ad}}} \\
                \hline
                SVD & $\Tensor{T}_{64,48}^{32,24}$ & $\Tensor{T}_{64,48}^{32,24}$ & $\Tensor{T}_{64,48}^{32,24}$ & $\Tensor{T}_{64,48}^{32,24}$\\
                OPDF (Ours) & $\Tensor{T}_{64,1,1,1,1,1,1,1,1,48}^{32,1,1,1,1,1,1,1,1,24}$ & $\Tensor{T}_{64,1,1,1,48}^{32,1,1,1,24}$ & $\Tensor{T}_{64,1,1,1,1,1,1,48}^{32,1,1,1,1,1,1,24}$ & 
                $\Tensor{T}_{64,1, 48}^{32,1, 24}$\\
                \hline
                \hline
                Experiments & \thead{ SST-2} &  \thead{ QNLI} & \thead{ QQP} &  \thead{ MNLI} \\
                \hline
                \rowcolor{gray!10}
                \multicolumn{5}{c}{\textbf{BERT-of-Theseus}} \\
                \hline
                SVD & $\Tensor{T}_{64,48}^{32,24}$ & $\Tensor{T}_{64,48}^{32,24}$ & $\Tensor{T}_{64,48}^{32,24}$ & $\Tensor{T}_{64,48}^{32,24}$\\
                OPDF (Ours) & $\Tensor{T}_{64,1,1,1,1,1,48}^{32,1,1,1,1,1,24}$ & $\Tensor{T}_{64,1,1,1,1,1,1,1,1,48}^{32,1,1,1,1,1,1,1,1,24}$ & $\Tensor{T}_{64,1,1,1,48}^{32,1,1,1,24}$ & $\Tensor{T}_{64,1,48}^{32,1,24}$\\
                \hline
                \rowcolor{gray!10}
                \multicolumn{5}{c}{\textbf{LGTM}} \\
                \hline
                SVD & $\Tensor{T}_{64,48}^{32,24}$ & $\Tensor{T}_{64,48}^{32,24}$ & $\Tensor{T}_{64,48}^{32,24}$ & $\Tensor{T}_{64,48}^{32,24}$\\
                OPDF (Ours) & $\Tensor{T}_{64,1,1,1,48}^{32,1,1,1,24}$ & $\Tensor{T}_{64,1,1,1,48}^{32,1,1,1,24}$ & $\Tensor{T}_{64,1,48}^{32,1,24}$ & $\Tensor{T}_{64,1,1,1,48}^{32,1,1,1,24}$\\
                \hline
                \rowcolor{gray!10}
                \multicolumn{5}{c}{\textbf{DBKD}} \\
                \hline
                SVD & $\Tensor{T}_{64,48}^{32,24}$ & $\Tensor{T}_{64,48}^{32,24}$ & $\Tensor{T}_{64,48}^{32,24}$ & $\Tensor{T}_{64,48}^{32,24}$\\
                OPDF (Ours) & $\Tensor{T}_{64,1,1,1,1,1,48}^{32,1,1,1,1,1,24}$ &
                $\Tensor{T}_{64,1,1,1,1,1,1,1,1,48}^{32,1,1,1,1,1,1,1,1,24}$ & 
                $\Tensor{T}_{64,1,1,1,1,48}^{32,1,1,1,1,24}$ & 
                $\Tensor{T}_{64,1,1,1,1,1,1,1,1,48}^{32,1,1,1,1,1,1,1,1,24}$\\
                \hline
                \rowcolor{gray!10}
                \multicolumn{5}{c}{\textbf{AD-KD}} \\
                \hline
                SVD & $\Tensor{T}_{64,48}^{32,24}$ & $\Tensor{T}_{64,48}^{32,24}$ & $\Tensor{T}_{64,48}^{32,24}$ & $\Tensor{T}_{64,48}^{32,24}$\\
                OPDF (Ours) & $\Tensor{T}_{64,1,1,1,48}^{32,1,1,1,24}$ & $\Tensor{T}_{64,1,1,1,48}^{32,1,1,1,24}$ & $\Tensor{T}_{64,1,1,1,48}^{32,1,1,1,24}$ & $\Tensor{T}_{64,1,1,1,1,48}^{32,1,1,1,1,24}$\\
                \hline
                \end{tabular}
        }

    \label{tab:hyper_nlp_ffn}
\end{table*}

\begin{table*}[h]
    \centering
    \caption{The multi-head attention layer settings in NLP distilation model.}
        \resizebox{\columnwidth}{!}{
            \begin{tabular}{l|cccc}
                \hline
                Experiments & \thead{ RTE} &  \thead{ MRPC} &  \thead{ STS-B} &  \thead{ CoLA}\\
                \hline
                \rowcolor{gray!10}
                \multicolumn{5}{c}{\textbf{BERT-of-Theseus}} \\
                \hline
                SVD & $\Tensor{T}_{32,24}^{32,24}$ & $\Tensor{T}_{32,24}^{32,24}$ & $\Tensor{T}_{32,24}^{32,24}$ & $\Tensor{T}_{32,24}^{32,24}$\\
                OPDF (Ours) & $\Tensor{T}_{32,1,1,1,24}^{32,1,1,1,24}$ & $\Tensor{T}_{32,1,1,1,24}^{32,1,1,1,24}$ & $\Tensor{T}_{32,1,1,1,24}^{32,1,1,1,24}$ & $\Tensor{T}_{32,1,24}^{32,1,24}$\\
                \hline
                \rowcolor{gray!10}
                \multicolumn{5}{c}{\textbf{LGTM}} \\
                \hline
                SVD & $\Tensor{T}_{32,24}^{32,24}$ & $\Tensor{T}_{32,24}^{32,24}$ & $\Tensor{T}_{32,24}^{32,24}$ & $\Tensor{T}_{32,24}^{32,24}$\\
                OPDF (Ours) & $\Tensor{T}_{32,1,1,1,24}^{32,1,1,1,24}$ & $\Tensor{T}_{32,1,1,1,24}^{32,1,1,1,24}$ & $\Tensor{T}_{32,1,1,1,24}^{32,1,1,1,24}$ & $\Tensor{T}_{32,1,1,1,1,24}^{32,1,1,1,1,24}$\\
                \hline
                \rowcolor{gray!10}
                \multicolumn{5}{c}{\textbf{DBKD}} \\
                \hline
                SVD & $\Tensor{T}_{32,24}^{32,24}$ & $\Tensor{T}_{32,24}^{32,24}$ & / & $\Tensor{T}_{32,24}^{32,24}$\\
                OPDF (Ours) & $\Tensor{T}_{32,1,1,1,1,24}^{32,1,1,1,1,24}$ &
                $\Tensor{T}_{32,1,1,1,1,1,1,1,1,24}^{32,1,1,1,1,1,1,1,1,24}$ & 
                / & 
                $\Tensor{T}_{32,1,1,1,1,1,1,1,24}^{32,1,1,1,1,1,1,1,24}$\\
                \hline
                \rowcolor{gray!10}
                \multicolumn{5}{c}{\textbf{AD-KD}} \\
                \hline
                SVD & $\Tensor{T}_{32,24}^{32,24}$ & $\Tensor{T}_{32,24}^{32,24}$ & $\Tensor{T}_{32,24}^{32,24}$ & $\Tensor{T}_{32,24}^{32,24}$\\
                OPDF (Ours) & $\Tensor{T}_{32,1,1,1,1,1,1,1,1,24}^{32,1,1,1,1,1,1,1,1,24}$ & $\Tensor{T}_{32,1,1,1,24}^{32,1,1,1,24}$ & $\Tensor{T}_{32,1,1,1,1,1,1,24}^{32,1,1,1,1,1,1,24}$ & 
                $\Tensor{T}_{32,1,24}^{32,1,24}$\\
                \hline              
                \hline
                Experiments &  \thead{ SST-2} &  \thead{ QNLI} & \thead{ QQP} &  \thead{ MNLI} \\
                \hline
                \rowcolor{gray!10}
                \multicolumn{5}{c}{\textbf{BERT-of-Theseus}} \\
                \hline
                SVD & $\Tensor{T}_{32,24}^{32,24}$ & $\Tensor{T}_{32,24}^{32,24}$ & $\Tensor{T}_{32,24}^{32,24}$ & $\Tensor{T}_{32,24}^{32,24}$\\
                OPDF (Ours) & $\Tensor{T}_{32,1,1,1,1,1,24}^{32,1,1,1,1,1,24}$ & $\Tensor{T}_{32,1,1,1,1,1,1,1,1,24}^{32,1,1,1,1,1,1,1,1,24}$ & $\Tensor{T}_{32,1,1,1,24}^{32,1,1,1,24}$ & $\Tensor{T}_{32,1,24}^{32,1,24}$\\
                \hline
                \rowcolor{gray!10}
                \multicolumn{5}{c}{\textbf{LGTM}} \\
                \hline
                SVD & $\Tensor{T}_{32,24}^{32,24}$ & $\Tensor{T}_{32,24}^{32,24}$ & $\Tensor{T}_{32,24}^{32,24}$ & $\Tensor{T}_{32,24}^{32,24}$\\
                OPDF (Ours) & $\Tensor{T}_{32,1,1,1,24}^{32,1,1,1,24}$ & $\Tensor{T}_{32,1,1,1,24}^{32,1,1,1,24}$ & $\Tensor{T}_{32,1,24}^{32,1,24}$ & $\Tensor{T}_{32,1,1,1,24}^{32,1,1,1,24}$\\
                \hline
                \rowcolor{gray!10}
                \multicolumn{5}{c}{\textbf{DBKD}} \\
                \hline
                SVD & $\Tensor{T}_{32,24}^{32,24}$ & $\Tensor{T}_{32,24}^{32,24}$ & $\Tensor{T}_{32,24}^{32,24}$ & $\Tensor{T}_{32,24}^{32,24}$\\
                OPDF (Ours) & $\Tensor{T}_{32,1,1,1,1,1,24}^{32,1,1,1,1,1,24}$ &
                $\Tensor{T}_{32,1,1,1,1,1,1,1,1,24}^{32,1,1,1,1,1,1,1,1,24}$ & 
                $\Tensor{T}_{32,1,1,1,1,24}^{32,1,1,1,1,24}$ & 
                $\Tensor{T}_{32,1,1,1,1,1,1,1,1,24}^{32,1,1,1,1,1,1,1,1,24}$\\
                \hline
                \rowcolor{gray!10}
                \multicolumn{5}{c}{\textbf{AD-KD}} \\
                \hline
                SVD & $\Tensor{T}_{32,24}^{32,24}$ & $\Tensor{T}_{32,24}^{32,24}$ & $\Tensor{T}_{32,24}^{32,24}$ & $\Tensor{T}_{32,24}^{32,24}$\\
                OPDF (Ours) & $\Tensor{T}_{32,1,1,1,24}^{32,1,1,1,24}$ & $\Tensor{T}_{32,1,1,1,24}^{32,1,1,1,24}$ & $\Tensor{T}_{32,1,1,1,24}^{32,1,1,1,24}$ & $\Tensor{T}_{32,1,1,1,1,24}^{32,1,1,1,1,24}$\\
                \hline
                \end{tabular}
        }

    \label{tab:hyper_nlp_att}
\end{table*}

% We also apply OPDF on TinyViT~\cite{wu2022tinyvit} to show it is an orthogonal method for various KD methods based on the Transformer. The experimental parameter configuration is shown in Table\ref{tab:hyper_cv}.

Additionally, we implement OPDF on TinyViT~\cite{wu2022tinyvit} to demonstrate its applicability as an orthogonal approach across various knowledge distillation methods that utilize the transformer architecture. Unlike NLP models, we decompose the projection layer in addition to the feed-forward network and multi-head attention layer in the vision transformer block. The specific experimental parameters utilized are detailed in Table \ref{tab:hyper_cv}. In CV tasks, our method takes 160.0 GPU days on A100 GPUs to pretrain TinyViT-21M.
\begin{table*}[h]

    \centering
    % \caption{The linear probe performance on imagenet-1k~\cite{deng2009imagenet}, imagenet Real~\cite{beyer2020we}, imagenet v2~\cite{recht2019imagenet} of TinyVit\cite{wu2022tinyvit} pretrained on imagenet-21k(in percent). Numbers marked with * indicate that these results are got by official checkpoint and released code. The best result for each task is highlight in bold. "\#Train Params" and "\#Inference Params" respectively denote the number of total parameters during training and inference. }
    \caption{The experiment settings in CV distilation model.}
        \resizebox{\columnwidth}{!}{
            \begin{tabular}{l|ccc|ccc|ccc}
                \hline
                \multirow{2}{*}{\centering Experiments} & \multicolumn{3}{c|}{Feed-forward Network} &  \multicolumn{3}{c|}{Multi-head Attention} &  \multicolumn{3}{c}{Projection Layer} \\
                  & layer1 &  layer2 &  layer3 & layer1 &  layer2 &  layer3 & layer1 &  layer2 &  layer3  \\
                \hline
                \rowcolor{gray!10}
                \multicolumn{10}{c}{\textbf{TinyVit-5M}} \\
                \hline
                SVD & $\Tensor{T}_{32,16}^{16,8}$ & $\Tensor{T}_{32,20}^{16,10}$ & $\Tensor{T}_{40,32}^{20,16}$ 
                & $\Tensor{T}_{24,16}^{16,8}$ & $\Tensor{T}_{24,20}^{16,10}$ & $\Tensor{T}_{32,30}^{20,16}$ 
                & $\Tensor{T}_{16,8}^{16,8}$ & $\Tensor{T}_{16,10}^{16,10}$ & $\Tensor{T}_{20,16}^{20,16}$ \\
                OPDF (Ours) & $\Tensor{T}_{32,1,16}^{16,1,8}$ & $\Tensor{T}_{32,1,20}^{16,1,10}$ & $\Tensor{T}_{40,1,32}^{20,1,16}$ 
                & $\Tensor{T}_{24,1,16}^{16,8}$ & $\Tensor{T}_{24,1,20}^{16,1,10}$ & $\Tensor{T}_{32,1,30}^{20,1,16}$ 
                & $\Tensor{T}_{16,1,8}^{16,1,8}$ & $\Tensor{T}_{16,1,10}^{16,1,10}$ & $\Tensor{T}_{20,1,16}^{20,1,16}$ \\

                \hline
                \rowcolor{gray!10}
                \multicolumn{10}{c}{\textbf{TinyVit-11M}} \\
                \hline
                SVD & $\Tensor{T}_{32,16}^{16,8}$ & $\Tensor{T}_{32,32}^{16,16}$ & $\Tensor{T}_{56,32}^{32,14}$ 
                & $\Tensor{T}_{24,16}^{16,8}$ & $\Tensor{T}_{32,24}^{16,16}$ & $\Tensor{T}_{48,28}^{32,14}$ 
                & $\Tensor{T}_{16,8}^{16,8}$ & $\Tensor{T}_{16,16}^{16,16}$ & $\Tensor{T}_{32,14}^{32,14}$ \\
                OPDF (Ours) & $\Tensor{T}_{32,1,16}^{16,1,8}$ & $\Tensor{T}_{32,1,32}^{16,1,16}$ & $\Tensor{T}_{56,1,32}^{32,1,14}$ 
                & $\Tensor{T}_{24,1,16}^{16,1,8}$ & $\Tensor{T}_{32,1,24}^{16,1,16}$ & $\Tensor{T}_{48,1,28}^{32,1,14}$ 
                & $\Tensor{T}_{16,1,8}^{16,1,8}$ & $\Tensor{T}_{16,1,16}^{16,1,16}$ & $\Tensor{T}_{32,1,14}^{32,1,14}$ \\

                \hline
                \rowcolor{gray!10}
                \multicolumn{10}{c}{\textbf{TinyVit-21M}} \\
                \hline
                SVD & $\Tensor{T}_{32,24}^{24,8}$ & $\Tensor{T}_{48,32}^{24,16}$ & $\Tensor{T}_{64,36}^{32,18}$ 
                & $\Tensor{T}_{32,18}^{24,8}$ & $\Tensor{T}_{36,32}^{24,16}$ & $\Tensor{T}_{54,32}^{32,18}$ 
                & $\Tensor{T}_{24,8}^{24,8}$ & $\Tensor{T}_{24,16}^{24,16}$ & $\Tensor{T}_{32,18}^{32,18}$ \\
                OPDF (Ours) & $\Tensor{T}_{32,1,24}^{24,1,8}$ & $\Tensor{T}_{48,1,32}^{24,1,16}$ & $\Tensor{T}_{64,1,36}^{32,1,18}$ 
                & $\Tensor{T}_{32,1,18}^{24,1,8}$ & $\Tensor{T}_{36,1,32}^{24,1,16}$ & $\Tensor{T}_{54,1,32}^{32,1,18}$ 
                & $\Tensor{T}_{24,1,8}^{24,1,8}$ & $\Tensor{T}_{24,1,16}^{24,1,16}$ & $\Tensor{T}_{32,1,18}^{32,1,18}$ \\
                \hline
                \end{tabular}
        }

    \label{tab:hyper_cv}
\end{table*}

We report the performance of the model that achieves the best results on the validation set when applied to the test set.

% \subsection{Impact of MPO methods}
% \label{impact_of_mpo}

% To demonstrate the robustness of our MPO methods, we applied different MPO methods to the DBKD and AD-KD model on the RTE, MRPC, STS-B, CoLA, and SST-2 task. The experimental results are presented in Table~\ref{tab:mpo_method}. To maintain consistent over-parameterization scales, we used the same decomposition scale~(L) for each KD model across the same task.
% \input{section/table/mpo_methods}
% We can observe that the performance of our approach consistently stabilizes around certain values, indicating that our method is not sensitive to the specific MPO techniques used. Therefore, when over-parameterizing, we should focus primarily on the decomposition scale rather than the MPO method employed.

%%%%%%%%%%%%%%%%%%%%%%%%%%%%%%%%%%%%%%%%%%%%%%%%%%%%%%%%%%%%
\clearpage
\newpage
\section*{NeurIPS Paper Checklist}

\begin{enumerate}

\item {\bf Claims}
    \item[] Question: Do the main claims made in the abstract and introduction accurately reflect the paper's contributions and scope?
    \item[] Answer: \answerYes{} % Replace by \answerYes{}, \answerNo{}, or \answerNA{}.
    \item[] Justification: Detailed descriptions of all main claims can be found in Section~\ref{method}. Furthermore, the main claims made in the abstract and introduction are supported by the experimental results presented in Section~\ref{experiments}.
    \item[] Guidelines:
    \begin{itemize}
        \item The answer NA means that the abstract and introduction do not include the claims made in the paper.
        \item The abstract and/or introduction should clearly state the claims made, including the contributions made in the paper and important assumptions and limitations. A No or NA answer to this question will not be perceived well by the reviewers. 
        \item The claims made should match theoretical and experimental results, and reflect how much the results can be expected to generalize to other settings. 
        \item It is fine to include aspirational goals as motivation as long as it is clear that these goals are not attained by the paper. 
    \end{itemize}

\item {\bf Limitations}
    \item[] Question: Does the paper discuss the limitations of the work performed by the authors?
    \item[] Answer: \answerYes{} % Replace by \answerYes{}, \answerNo{}, or \answerNA{}.
    \item[] Justification: See Section~\ref{scale}. There are inherent limits to the benefits that can be achieved through over-parameterization in knowledge distillation models. The learning rate should be chosen more carefully as the scale of over-parameterization increases. % 学习率需要精细调整，而且上限有限制
    \item[] Guidelines:
    \begin{itemize}
        \item The answer NA means that the paper has no limitation while the answer No means that the paper has limitations, but those are not discussed in the paper. 
        \item The authors are encouraged to create a separate "Limitations" section in their paper.
        \item The paper should point out any strong assumptions and how robust the results are to violations of these assumptions (e.g., independence assumptions, noiseless settings, model well-specification, asymptotic approximations only holding locally). The authors should reflect on how these assumptions might be violated in practice and what the implications would be.
        \item The authors should reflect on the scope of the claims made, e.g., if the approach was only tested on a few datasets or with a few runs. In general, empirical results often depend on implicit assumptions, which should be articulated.
        \item The authors should reflect on the factors that influence the performance of the approach. For example, a facial recognition algorithm may perform poorly when image resolution is low or images are taken in low lighting. Or a speech-to-text system might not be used reliably to provide closed captions for online lectures because it fails to handle technical jargon.
        \item The authors should discuss the computational efficiency of the proposed algorithms and how they scale with dataset size.
        \item If applicable, the authors should discuss possible limitations of their approach to address problems of privacy and fairness.
        \item While the authors might fear that complete honesty about limitations might be used by reviewers as grounds for rejection, a worse outcome might be that reviewers discover limitations that aren't acknowledged in the paper. The authors should use their best judgment and recognize that individual actions in favor of transparency play an important role in developing norms that preserve the integrity of the community. Reviewers will be specifically instructed to not penalize honesty concerning limitations.
    \end{itemize}

\item {\bf Theory Assumptions and Proofs}
    \item[] Question: For each theoretical result, does the paper provide the full set of assumptions and a complete (and correct) proof?
    \item[] Answer: \answerYes{} % Replace by \answerYes{}, \answerNo{}, or \answerNA{}.
    \item[] Justification: See Appendix~\ref{appendix_a} and~\ref{appendix_b}% 附录
    \item[] Guidelines:
    \begin{itemize}
        \item The answer NA means that the paper does not include theoretical results. 
        \item All the theorems, formulas, and proofs in the paper should be numbered and cross-referenced.
        \item All assumptions should be clearly stated or referenced in the statement of any theorems.
        \item The proofs can either appear in the main paper or the supplemental material, but if they appear in the supplemental material, the authors are encouraged to provide a short proof sketch to provide intuition. 
        \item Inversely, any informal proof provided in the core of the paper should be complemented by formal proofs provided in appendix or supplemental material.
        \item Theorems and Lemmas that the proof relies upon should be properly referenced. 
    \end{itemize}

    \item {\bf Experimental Result Reproducibility}
    \item[] Question: Does the paper fully disclose all the information needed to reproduce the main experimental results of the paper to the extent that it affects the main claims and/or conclusions of the paper (regardless of whether the code and data are provided or not)?
    \item[] Answer: \answerYes{} % Replace by \answerYes{}, \answerNo{}, or \answerNA{}.
    \item[] Justification: The framework of OPDF is detailed in Section~\ref{method}, and the experimental details are provided in Appendix~\ref{experimental_details}.
    \item[] Guidelines:
    \begin{itemize}
        \item The answer NA means that the paper does not include experiments.
        \item If the paper includes experiments, a No answer to this question will not be perceived well by the reviewers: Making the paper reproducible is important, regardless of whether the code and data are provided or not.
        \item If the contribution is a dataset and/or model, the authors should describe the steps taken to make their results reproducible or verifiable. 
        \item Depending on the contribution, reproducibility can be accomplished in various ways. For example, if the contribution is a novel architecture, describing the architecture fully might suffice, or if the contribution is a specific model and empirical evaluation, it may be necessary to either make it possible for others to replicate the model with the same dataset, or provide access to the model. In general. releasing code and data is often one good way to accomplish this, but reproducibility can also be provided via detailed instructions for how to replicate the results, access to a hosted model (e.g., in the case of a large language model), releasing of a model checkpoint, or other means that are appropriate to the research performed.
        \item While NeurIPS does not require releasing code, the conference does require all submissions to provide some reasonable avenue for reproducibility, which may depend on the nature of the contribution. For example
        \begin{enumerate}
            \item If the contribution is primarily a new algorithm, the paper should make it clear how to reproduce that algorithm.
            \item If the contribution is primarily a new model architecture, the paper should describe the architecture clearly and fully.
            \item If the contribution is a new model (e.g., a large language model), then there should either be a way to access this model for reproducing the results or a way to reproduce the model (e.g., with an open-source dataset or instructions for how to construct the dataset).
            \item We recognize that reproducibility may be tricky in some cases, in which case authors are welcome to describe the particular way they provide for reproducibility. In the case of closed-source models, it may be that access to the model is limited in some way (e.g., to registered users), but it should be possible for other researchers to have some path to reproducing or verifying the results.
        \end{enumerate}
    \end{itemize}

\item {\bf Open access to data and code}
    \item[] Question: Does the paper provide open access to the data and code, with sufficient instructions to faithfully reproduce the main experimental results, as described in supplemental material?
    \item[] Answer: \answerYes{} % Replace by \answerYes{}, \answerNo{}, or \answerNA{}.
    \item[] Justification: Our code is available at~\url{https://github.com/intell-sci-comput/OPDF}.
    \item[] Guidelines:
    \begin{itemize}
        \item The answer NA means that paper does not include experiments requiring code.
        \item Please see the NeurIPS code and data submission guidelines (\url{https://nips.cc/public/guides/CodeSubmissionPolicy}) for more details.
        \item While we encourage the release of code and data, we understand that this might not be possible, so “No” is an acceptable answer. Papers cannot be rejected simply for not including code, unless this is central to the contribution (e.g., for a new open-source benchmark).
        \item The instructions should contain the exact command and environment needed to run to reproduce the results. See the NeurIPS code and data submission guidelines (\url{https://nips.cc/public/guides/CodeSubmissionPolicy}) for more details.
        \item The authors should provide instructions on data access and preparation, including how to access the raw data, preprocessed data, intermediate data, and generated data, etc.
        \item The authors should provide scripts to reproduce all experimental results for the new proposed method and baselines. If only a subset of experiments are reproducible, they should state which ones are omitted from the script and why.
        \item At submission time, to preserve anonymity, the authors should release anonymized versions (if applicable).
        \item Providing as much information as possible in supplemental material (appended to the paper) is recommended, but including URLs to data and code is permitted.
    \end{itemize}

\item {\bf Experimental Setting/Details}
    \item[] Question: Does the paper specify all the training and test details (e.g., data splits, hyperparameters, how they were chosen, type of optimizer, etc.) necessary to understand the results?
    \item[] Answer:  \answerYes{} % Replace by \answerYes{}, \answerNo{}, or \answerNA{}.
    \item[] Justification: See Appendix~\ref{experimental_details}.
    \item[] Guidelines:
    \begin{itemize}
        \item The answer NA means that the paper does not include experiments.
        \item The experimental setting should be presented in the core of the paper to a level of detail that is necessary to appreciate the results and make sense of them.
        \item The full details can be provided either with the code, in appendix, or as supplemental material.
    \end{itemize}

\item {\bf Experiment Statistical Significance}
    \item[] Question: Does the paper report error bars suitably and correctly defined or other appropriate information about the statistical significance of the experiments?
    \item[] Answer: \answerYes{} % Replace by \answerYes{}, \answerNo{}, or \answerNA{}.
    \item[] Justification: For all the results, we report the mean values of five runs using different random seeds.
    \item[] Guidelines:
    \begin{itemize}
        \item The answer NA means that the paper does not include experiments.
        \item The authors should answer "Yes" if the results are accompanied by error bars, confidence intervals, or statistical significance tests, at least for the experiments that support the main claims of the paper.
        \item The factors of variability that the error bars are capturing should be clearly stated (for example, train/test split, initialization, random drawing of some parameter, or overall run with given experimental conditions).
        \item The method for calculating the error bars should be explained (closed form formula, call to a library function, bootstrap, etc.)
        \item The assumptions made should be given (e.g., Normally distributed errors).
        \item It should be clear whether the error bar is the standard deviation or the standard error of the mean.
        \item It is OK to report 1-sigma error bars, but one should state it. The authors should preferably report a 2-sigma error bar than state that they have a 96\% CI, if the hypothesis of Normality of errors is not verified.
        \item For asymmetric distributions, the authors should be careful not to show in tables or figures symmetric error bars that would yield results that are out of range (e.g. negative error rates).
        \item If error bars are reported in tables or plots, The authors should explain in the text how they were calculated and reference the corresponding figures or tables in the text.
    \end{itemize}

\item {\bf Experiments Compute Resources}
    \item[] Question: For each experiment, does the paper provide sufficient information on the computer resources (type of compute workers, memory, time of execution) needed to reproduce the experiments?
    \item[] Answer:  \answerYes{} % Replace by \answerYes{}, \answerNo{}, or \answerNA{}.
    \item[] Justification: See Appendix~\ref{experimental_details}.
    \item[] Guidelines:
    \begin{itemize}
        \item The answer NA means that the paper does not include experiments.
        \item The paper should indicate the type of compute workers CPU or GPU, internal cluster, or cloud provider, including relevant memory and storage.
        \item The paper should provide the amount of compute required for each of the individual experimental runs as well as estimate the total compute. 
        \item The paper should disclose whether the full research project required more compute than the experiments reported in the paper (e.g., preliminary or failed experiments that didn't make it into the paper). 
    \end{itemize}
    
\item {\bf Code Of Ethics}
    \item[] Question: Does the research conducted in the paper conform, in every respect, with the NeurIPS Code of Ethics \url{https://neurips.cc/public/EthicsGuidelines}?
    \item[] Answer: \answerYes{} % Replace by \answerYes{}, \answerNo{}, or \answerNA{}.
    \item[] Justification: See Section~\ref{impacts}
    \item[] Guidelines:
    \begin{itemize}
        \item The answer NA means that the authors have not reviewed the NeurIPS Code of Ethics.
        \item If the authors answer No, they should explain the special circumstances that require a deviation from the Code of Ethics.
        \item The authors should make sure to preserve anonymity (e.g., if there is a special consideration due to laws or regulations in their jurisdiction).
    \end{itemize}

\item {\bf Broader Impacts}
    \item[] Question: Does the paper discuss both potential positive societal impacts and negative societal impacts of the work performed?
    \item[] Answer: \answerYes{} % Replace by \answerYes{}, \answerNo{}, or \answerNA{}.
    \item[] Justification: See Section~\ref{impacts}
    \item[] Guidelines:
    \begin{itemize}
        \item The answer NA means that there is no societal impact of the work performed.
        \item If the authors answer NA or No, they should explain why their work has no societal impact or why the paper does not address societal impact.
        \item Examples of negative societal impacts include potential malicious or unintended uses (e.g., disinformation, generating fake profiles, surveillance), fairness considerations (e.g., deployment of technologies that could make decisions that unfairly impact specific groups), privacy considerations, and security considerations.
        \item The conference expects that many papers will be foundational research and not tied to particular applications, let alone deployments. However, if there is a direct path to any negative applications, the authors should point it out. For example, it is legitimate to point out that an improvement in the quality of generative models could be used to generate deepfakes for disinformation. On the other hand, it is not needed to point out that a generic algorithm for optimizing neural networks could enable people to train models that generate Deepfakes faster.
        \item The authors should consider possible harms that could arise when the technology is being used as intended and functioning correctly, harms that could arise when the technology is being used as intended but gives incorrect results, and harms following from (intentional or unintentional) misuse of the technology.
        \item If there are negative societal impacts, the authors could also discuss possible mitigation strategies (e.g., gated release of models, providing defenses in addition to attacks, mechanisms for monitoring misuse, mechanisms to monitor how a system learns from feedback over time, improving the efficiency and accessibility of ML).
    \end{itemize}
    
\item {\bf Safeguards}
    \item[] Question: Does the paper describe safeguards that have been put in place for responsible release of data or models that have a high risk for misuse (e.g., pretrained language models, image generators, or scraped datasets)?
    \item[] Answer: \answerNA{} % Replace by \answerYes{}, \answerNo{}, or \answerNA{}.
    \item[] Justification: Guidelines: The paper poses no such risks.
    \item[] 
    \begin{itemize}
        \item The answer NA means that the paper poses no such risks.
        \item Released models that have a high risk for misuse or dual-use should be released with necessary safeguards to allow for controlled use of the model, for example by requiring that users adhere to usage guidelines or restrictions to access the model or implementing safety filters. 
        \item Datasets that have been scraped from the Internet could pose safety risks. The authors should describe how they avoided releasing unsafe images.
        \item We recognize that providing effective safeguards is challenging, and many papers do not require this, but we encourage authors to take this into account and make a best faith effort.
    \end{itemize}

\item {\bf Licenses for existing assets}
    \item[] Question: Are the creators or original owners of assets (e.g., code, data, models), used in the paper, properly credited and are the license and terms of use explicitly mentioned and properly respected?
    \item[] Answer: \answerYes{} % Replace by \answerYes{}, \answerNo{}, or \answerNA{}.
    \item[] Justification: We cite the original paper that produced the code package or dataset.
    \item[] Guidelines: 
    \begin{itemize}
        \item The answer NA means that the paper does not use existing assets.
        \item The authors should cite the original paper that produced the code package or dataset.
        \item The authors should state which version of the asset is used and, if possible, include a URL.
        \item The name of the license (e.g., CC-BY 4.0) should be included for each asset.
        \item For scraped data from a particular source (e.g., website), the copyright and terms of service of that source should be provided.
        \item If assets are released, the license, copyright information, and terms of use in the package should be provided. For popular datasets, \url{paperswithcode.com/datasets} has curated licenses for some datasets. Their licensing guide can help determine the license of a dataset.
        \item For existing datasets that are re-packaged, both the original license and the license of the derived asset (if it has changed) should be provided.
        \item If this information is not available online, the authors are encouraged to reach out to the asset's creators.
    \end{itemize}

\item {\bf New Assets}
    \item[] Question: Are new assets introduced in the paper well documented and is the documentation provided alongside the assets?
    \item[] Answer: \answerNA{} % Replace by \answerYes{}, \answerNo{}, or \answerNA{}.
    \item[] Justification: The paper does not release new assets.
    \item[] Guidelines:
    \begin{itemize}
        \item The answer NA means that the paper does not release new assets.
        \item Researchers should communicate the details of the dataset/code/model as part of their submissions via structured templates. This includes details about training, license, limitations, etc. 
        \item The paper should discuss whether and how consent was obtained from people whose asset is used.
        \item At submission time, remember to anonymize your assets (if applicable). You can either create an anonymized URL or include an anonymized zip file.
    \end{itemize}

\item {\bf Crowdsourcing and Research with Human Subjects}
    \item[] Question: For crowdsourcing experiments and research with human subjects, does the paper include the full text of instructions given to participants and screenshots, if applicable, as well as details about compensation (if any)? 
    \item[] Answer: \answerNA{} % Replace by \answerYes{}, \answerNo{}, or \answerNA{}.
    \item[] Justification: The paper does not involve crowdsourcing nor research with human subjects.
    \item[] Guidelines:
    \begin{itemize}
        \item The answer NA means that the paper does not involve crowdsourcing nor research with human subjects.
        \item Including this information in the supplemental material is fine, but if the main contribution of the paper involves human subjects, then as much detail as possible should be included in the main paper. 
        \item According to the NeurIPS Code of Ethics, workers involved in data collection, curation, or other labor should be paid at least the minimum wage in the country of the data collector. 
    \end{itemize}

\item {\bf Institutional Review Board (IRB) Approvals or Equivalent for Research with Human Subjects}
    \item[] Question: Does the paper describe potential risks incurred by study participants, whether such risks were disclosed to the subjects, and whether Institutional Review Board (IRB) approvals (or an equivalent approval/review based on the requirements of your country or institution) were obtained?
    \item[] Answer: \answerNA{} % Replace by \answerYes{}, \answerNo{}, or \answerNA{}.
    \item[] Justification: The paper does not involve crowdsourcing nor research with human subjects.
    \item[] Guidelines:
    \begin{itemize}
        \item The answer NA means that the paper does not involve crowdsourcing nor research with human subjects.
        \item Depending on the country in which research is conducted, IRB approval (or equivalent) may be required for any human subjects research. If you obtained IRB approval, you should clearly state this in the paper. 
        \item We recognize that the procedures for this may vary significantly between institutions and locations, and we expect authors to adhere to the NeurIPS Code of Ethics and the guidelines for their institution. 
        \item For initial submissions, do not include any information that would break anonymity (if applicable), such as the institution conducting the review.
    \end{itemize}

\end{enumerate}

\end{document}